\documentclass{article}
\usepackage{arxiv}
\usepackage{graphicx}
\usepackage{subcaption}
\usepackage{amsfonts}
\usepackage{amsmath}
\usepackage{amssymb}
\usepackage{physics}
\usepackage{url}

\usepackage{graphicx}

\usepackage{amsmath}
\usepackage{amsfonts}
\usepackage{amssymb}

\newcommand{\etal}{{\it et al.}}

\usepackage{xcolor}

\newcommand{\aref}[1]{App.\,~\ref{#1}}
\newcommand{\fref}[1]{Fig.\,~\ref{#1}}
\newcommand{\tref}[1]{Table\,~\ref{#1}}
\newcommand{\eref}[1]{Eq.\,~(\ref{#1})}

\newcommand{\sref}[1]{Sec.\!~\ref{#1}}

\newcommand{\cref}[1]{Ref.\,~\cite{#1}}

\newcommand{\apriori}{{\it{a priori}}}

\newcommand{\bs}{\mathsf{b}}

\newcommand{\Ds}{\mathsf{D}}

\newcommand{\Vs}{\mathsf{V}}
\newcommand{\Ws}{\mathsf{W}}
\newcommand{\Xs}{\mathsf{X}}

\newcommand{\Ys}{\mathsf{Y}}

\newcommand{\eb}{\mathbf{e}}

\newcommand{\Eb}{\mathbf{E}}

\newcommand{\Ib}{\mathbf{I}}

\newcommand{\Lc}{\mathcal{L}}

\newcommand{\Cbb}{\mathbb{C}}
\newcommand{\Ebb}{\mathbb{E}}
\newcommand{\Ibb}{\mathbb{I}}

\newcommand{\alphab}{\boldsymbol{\alpha}}

\newcommand{\partialb}{{\boldsymbol{\partial}}}

\newcommand{\argmin}{\operatorname{argmin}}

\newcommand{\NN}{\mathsf{N}\!\mathsf{N}}

\newcommand{\parameters}{{\boldsymbol{\theta}}}
\newcommand{\inputvector}{\mathsf{x}}
\newcommand{\outputvector}{\mathsf{y}}
\newcommand{\hiddenvector}{\mathsf{h}}

\newcommand{\prob}{\pi}

\newcommand{\data}{\mathsf{D}}

\newcommand{\stress}{\mathbf{S}}
\newcommand{\strain}{\mathbf{E}}

\newcommand{\loss}{\mathcal{L}}
\newcommand{\MSE}{\operatorname{MSE}}
\newcommand{\LSE}{\operatorname{LSE}}

\title{\bf Differentiable neural network representation of  multi-well, locally-convex potentials}
\renewcommand{\shorttitle}{\bf Neural network representation of multi-well potentials}
\author{
Reese E. Jones \\
Sandia National Laboratories,\\
Livermore, CA, USA
\And
Adrian Buganza Tepole \\
Columbia University \\
New York City, NY, USA
\And
Jan N. Fuhg \\
The University of Texas at Austin\\
Austin, TX, USA
}
\date{}
\begin{document}

\maketitle

\begin{abstract}

Multi-well potentials are ubiquitous in science, modeling phenomena such as phase transitions, dynamic instabilities, and multimodal behavior across physics, chemistry, and biology. In contrast to non-smooth minimum-of-mixture representations, we propose a differentiable and convex formulation based on a log-sum-exponential (LSE) mixture of input convex neural network (ICNN) modes. This log-sum-exponential input convex neural network (LSE-ICNN) provides a smooth surrogate that retains convexity within basins and allows for gradient-based learning and inference.

A key feature of the LSE-ICNN is its ability to automatically discover both the number of modes and the scale of transitions through sparse regression, enabling adaptive and parsimonious modeling. We demonstrate the versatility of the LSE-ICNN across diverse domains, including mechanochemical phase transformations, microstructural elastic instabilities, conservative biological gene circuits, and variational inference for multimodal probability distributions. These examples highlight the effectiveness of the LSE-ICNN in capturing complex multimodal landscapes while preserving differentiability, making it broadly applicable in data-driven modeling, optimization, and physical simulation.
\end{abstract}

\section{Introduction}

Multi‑well energy landscapes are ubiquitous in physics, chemistry, and engineering.
For example, they govern classical particle mechanics in isomerization \cite{fonseca1983classical,sen1996relaxation}, quantum tunneling in double‑well potentials \cite{harrell1980double,berezovoj2010multi,jelic2012double}, transition‑state kinetics~\cite{laidler1983development,truhlar1996current}, micro‑structural phase transformations in solids~\cite{cozzi2017nonlocal,cailleau1980double}, and the hysteretic response of architected metamaterials~\cite{rossi2024limit}.
In addition to physical processes, multi-well potentials are common in probabilistic inference.
For example, multi‑modal posterior distributions arise whenever competing hypotheses explain the same data, or in ill-posed inverse problems where there is no unique solution ~\cite{yao2022stacking,yang2016multimodal,sutter2021generalized}.
Across all these domains, a differentiable surrogate of the underlying potential (or probability distribution) is indispensable for gradient‑based simulation, optimization, and learning.

A common ansatz for multi‑well potentials can be expressed as the minimum of a finite collection of convex functions \cite{kohn1991relaxation,firoozye1993geometric,pagano1998solid,smyshlyaev1999relation}.
While such a point-wise minimum preserves convexity within each basin, the global function is continuous but not differentiable everywhere, as it exhibits kinks along the interfaces where multiple wells attain the same value.
These non‑differentiable regions impede gradient- and Hessian-based solvers in mechanics, break adjoint‐based sensitivity analysis, and preclude direct use in modern deep‑learning frameworks that rely on automatic differentiation.

A differentiable function that approximates a maximum appears in statistical mechanics \cite{beale1996statistical} as the \textit{log‑partition} function in the Helmholtz free energy $F$:
\begin{equation*}
F= F({E_i},T) = -k_BT\ln Z \quad \text{where} \quad Z=\sum_i
\exp{((k_BT)^{-1} E_i)} \ ,
\end{equation*}
which is the log sum of exponential energetic contributions $\{ E_i \}$ tempered by the absolute temperature $T$.
As a smooth approximation of a multivariate maximum, the log-sum-exp (LSE) function (also known as \emph{real softmax} or \emph{softplus})  is widely employed in machine learning, and its gradient, the \emph{softmax}  function, is essential in logistic regression tasks \cite{bishop2006pattern}.
As the name suggests, the \textit{softmax} replaces the \textit{max} function with a smooth differentiable approximation \cite{gao2017properties}.
The LSE function, like the \textit{softmax} activation function, also yields a smooth approximation to \textit{min} or \textit{max} functions \cite{blanchard2021accurately}.
Furthermore, the LSE function is smooth and convex \cite{boyd2004convex}.
Consequently, it has been widely used for regularization in convex optimization problems of non-smooth functions \cite{beck2012smoothing,xu2001smoothing}, in particular in regularizing the minimum of convex functions \cite{nesterov2005smooth}.

Convexity of a potential conveys a number of theoretical properties that lead to guarantees of solvability and stability \cite{bethuel1999variational}.
Among general representations of convex functions, the so-called \emph{input convex} neural network (ICNN) \cite{amos2017input} provides a simple, expressive, and extensible framework based on a densely connected neural network with skip connections, weight constraints, and convex activations.
It has seen wide-spread applications in mechanics \cite{tac2022data,chen2022polyconvex,as2022mechanics,xu2021learning,klein2022polyconvex,klein2023parametrized,kalina2024neural,fuhg2022learning,fuhg2024polyconvex,jadoon2025automated} and other fields \cite{makkuva2020optimal,chen2018optimal}.

Our contribution is to combine: (a) the LSE, (b) the ICNN, and (c) a sparse regression loss to create a differentiable representation capable of fitting multi-well potentials with different scales of transition between a learned number of locally convex wells.

In the next section (\sref{sec:related}) we put our contribution in the context of related work.
Then in \sref{sec:method} we present the differentiable mixture layer followed by illustrative examples.
In \sref{sec:results} we demonstrate the method on complex data ranging from mechanochemistry to materials science.
Finally, we conclude with a summary of results and avenues for future work in \sref{sec:conclusion}.

\section{Related work} \label{sec:related}

Finding arbitrary potentials to fit experimental data through machine learning architectures has gained interest in several areas of computational physics.
In most applications, potentials are expected to be at least locally convex.
Convexity in data-driven potentials can be enforced \apriori with input convex neural networks (ICNNs) \cite{amos2017input,fuhg2021physics}, and similar architectures \cite{sankaranarayanan2022cdinn,jadoon2025input} designed to ensure convexity with respect to their inputs.
In solid mechanics, global polyconvexity of strain energy potentials is desirable to guarantee the existence and uniqueness of boundary‑value problems \cite{ball1976convexity}.
However, when the underlying material contains, for example, phase‑transforming crystals or snap‑through instabilities, the effective energy develops multiple local minima~\cite{braides1994loss,rossi2024limit}.
A straightforward way to capture such behavior in a data-driven way is to model the energy as the point-wise minimum of several convex wells modeled via ICNNs~\cite{kumar2020assessment,thakolkaran2025experiment}.
As discussed, the problem with this approach is that the potential is non‑differentiable along the intersection of the multiple wells.
Also, physical transitions/saddle points are generally smooth with a physical scale appropriate for the physics of the transition.
Furthermore, the point-wise minimum approach requires explicit knowledge of the number of distinct wells.

An alternative strategy for modeling multi-well energy landscapes in physics-informed settings involves sparse system identification, where symbolic or interpretable models are inferred directly from data using sparsity-promoting priors~\cite{zhang2021discovering}.
While promising, these approaches may struggle to scale or generalize to high dimensions.

Our own previous work has leveraged ICNNs to model polyconvex strain energy functions, and introduced gating mechanisms to detect spatial or temporal regions of dissipation versus elastic response\cite{fuhg2022learning,jones2025attention}.
This framework preserved the necessary convexity in conservative regimes while allowing transitions to inelastic behaviors through learned internal-state dynamics.

Within the related context of multimodal probability density distributions, efforts to create mixture models such as Gaussian mixture models are common~\cite{yu2018density}, and several data-driven enhancements have been proposed in this regard~\cite{fernandez2011melm,liu2021primal}.
While GMMs and their machine learning variants offer smooth representations, they do not typically enforce convexity of the underlying components, which is an essential feature in many physical systems.

\section{Neural network architecture} \label{sec:method}

To represent multi-well potentials that are locally convex, we develop a neural network based on a log-sum-exponential (LSE)  mixing of input convex NNs.

\begin{figure}
\centering
\includegraphics[width=0.5\linewidth]{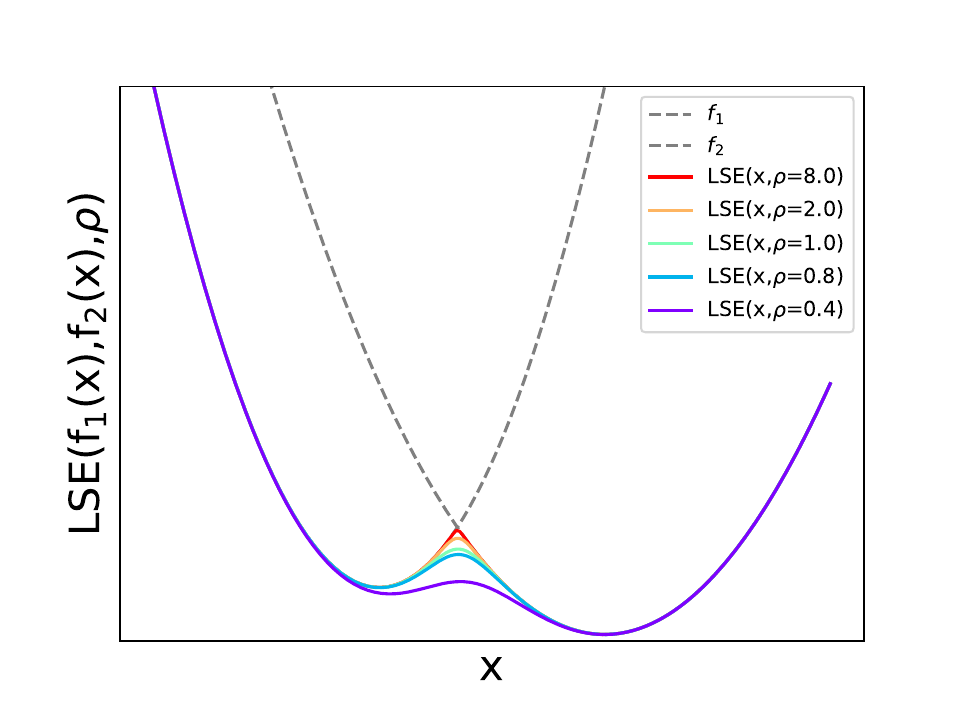}
\caption{Illustration of the effect of length-scale $\rho$ on the $\LSE$ function.}
\label{fig:rho}
\end{figure}

\paragraph{Log sum exponent layer}
Previous representations \cite{kumar2020assessment,thakolkaran2025experiment} rely on a simple minimum of a set of functions $\{ f_i(\inputvector) \}$
\begin{equation}
\min \left( f_{1}(\inputvector), \ldots, f_{n}(\inputvector) \right).
\end{equation}
We want a differentiable approximation of the minimum function with a learnable length-scale of transition.
The LSE function
\begin{equation} \label{eq:LSE}
\LSE(\{ f_i(\inputvector) \}, \rho) = - \frac{1}{\rho} \log\left(  \frac{1}{N_\text{modes}} \sum_{i=1}^{N_\text{modes}} \exp(-\rho f_{i}(\inputvector)) \right) \quad ,
\end{equation}
is smoothly differentiable and approaches $\min_{i} (f_{i})$ as $\rho \rightarrow +\infty$, see Boyd and Vandenberghe \cite{boyd2004convex} (Section 3.1.5).
This transition is illustrated in \fref{fig:rho}.
Here $\rho$ is analogous to the inverse temperature in the Helmholtz free energy.
To allow for the number of wells to be discoverable, we reframe \eref{eq:LSE} as a mode-weighted function:
\begin{equation} \label{eq:lse}
\LSE_{\alpha}(\{ f_i(\inputvector) \}, \rho) = - \frac{1}{\rho} \log\left( \frac{1}{N_\text{modes}} \sum_{i=1}^{N_\text{modes}} \varsigma(\alpha_{i}) \exp(-\rho f_{i}(\inputvector)) \right) \ ,
\end{equation}
where $\alpha_i$ are mode weights and $\varsigma$ is a gating activation function. In order to preserve convexity, we require that $\varsigma(\alpha_i)>0$ and $\rho>0$. Herein we employed a shifted and scaled sigmoid function for $\varsigma(x) = \operatorname{sigmoid}(10(x/2-1))$ such that $1>\varsigma(\alpha_i)>0$.
Alternatively, we can recast \eref{eq:lse} as
\begin{equation} \label{eq:lse2}
\begin{aligned}
\LSE_{\alpha}(\{ f_i(\inputvector) \}, \rho) =-\frac{1}{\rho} \log\left( \frac{1}{N_\text{modes}} \sum_{i=1}^{N_\text{modes}}   \exp(-\rho ( f_{i}(\inputvector) -\log(\varsigma(\alpha_{i})))) \right) \quad ,
\end{aligned}
\end{equation}
which further elucidates the role of the weighing function in the approximation of \eref{eq:LSE}.
To enable local convexity, i.e., convexity near distinct modes/wells, each $f_i$ in \eref{eq:lse} is an input convex neural network (ICNN).

\paragraph{Input convex neural network}
For each mode function $f_i(\inputvector)$, we employ an ICNN, which is a fully connected neural network with a skip connection and constraints on specific weights.
For an input vector $\inputvector$, the ICNN is a feedforward stack of $N_\ell$ layers:
\begin{eqnarray}
\hiddenvector_1 &=& \sigma_1 \left(
\Vs_1 \inputvector +\bs_1
\right) \nonumber \\
\hiddenvector_k &=& \sigma_k \left(
\Vs_k \inputvector + \Ws_k \hiddenvector_{k-1} +\bs_k
\right)  \qquad k=2, \ldots, N_\ell-1  \label{eq:ICNN} \\
\outputvector &=& \Vs_{N_\ell} \inputvector + \Ws_{N_\ell} \hiddenvector_{N-1} +\bs_{N_\ell} \nonumber
\end{eqnarray}
with weights $\Ws_k$ and $\Vs_k$, activation functions $\sigma_k$ and $k \in [1,N_\ell]$.
The weights and biases form the set of trainable parameters $\parameters = \lbrace \Ws_k,  \Vs_k, \bs_k \rbrace$.
The output $\outputvector = \NN(\inputvector;\parameters)$ is convex with respect to the input because the weights $\Ws_k$ are constrained to be non-negative, and the activation functions $\sigma_k$ are chosen to be convex and non-decreasing \cite{amos2017input} and set as softplus in this work.

\paragraph{Log-sum-exponential input convex neural network}
The LSE-ICNN is simply
\begin{equation} \label{eq:lse-icnn}
\LSE_{\alpha}(\{ \NN_i(\inputvector) \}, \rho)
= - \frac{1}{\rho} \log\left( \frac{1}{N_\text{modes}} \sum_{i=1}^{N_\text{modes}} \varsigma(\alpha_{i}) \exp(-\rho \, \NN_{i}(\inputvector)) \right) \ ,
\end{equation}
where $\rho$ is a learnable parameter in addition to the weights of the ICNN modes $\{ \NN_i \}$ and mode weights $\{ \alpha_i \}$.

\paragraph{Sparse regression}
The parameters of LSE-ICNN are $\parameters =  \{ \rho, \{ \alpha_i, \parameters_i \}_{i=1}^{N_\text{modes}} \}$.
We use the usual mean squared error (MSE) loss function augmented with an L1 penalty $\epsilon$ on the weights $\alphab = \{ \alpha_i \}$
\begin{equation}
\loss = \underbrace{\| \outputvector - \LSE(\inputvector; \parameters)\|_2}_{\MSE(\parameters; \data)}  + \varepsilon | \alphab |_1 \quad ,
\end{equation}
which is common in sparse regression \cite{bertsimas2020sparse} to minimize the number of non-zero components.
In practice, we start with a generous estimate of the number of modes $N_\text{modes}$ for the application and then, through the L1 regularization of $\alpha_{i}$ remove unnecessary terms during training with gradient descent.

\paragraph{Illustrations}
To illustrate the expressive capacity of the LSE-ICNN, we use it to fit one-dimensional functions using 200 randomly selected points chosen uniformly in  $\{x_i \in [-3,3]\}_i$ and plot the result over 200 equally spaced samples in the same range.
In particular, we fit to three multi-well functions: (a) an asymmetric quartic double well
\begin{equation}
y = \frac{2}{5}( \frac{1}{4} x^4 - x^2+ \frac{1}{8} x^3) \quad ,
\end{equation}
(b) a cosine-modulated quadratic well
\begin{equation}
y =  \frac{1}{20}\left( x^2 + 1.6 \cos(4 x) + 1 \right) \quad ,
\end{equation}
and (c) a 3-well minimum mixture
\begin{equation}
y = \min\left(\{ a_i f(x,x_i) + b_i \}_i \right) \quad ,
\end{equation}
where $f(x,x_i) = 0.05 (x-x_i)^2$ and
${x_i} = \{-1,    0, 2 \}$,
${a_i} = \{12,    8, 4 \}$,
${b_i} = \{ 2, -1/2, 1 \}$.

\begin{figure}[h]
\centering
\begin{subfigure}[c]{0.32\textwidth}
\includegraphics[width=0.99\linewidth]{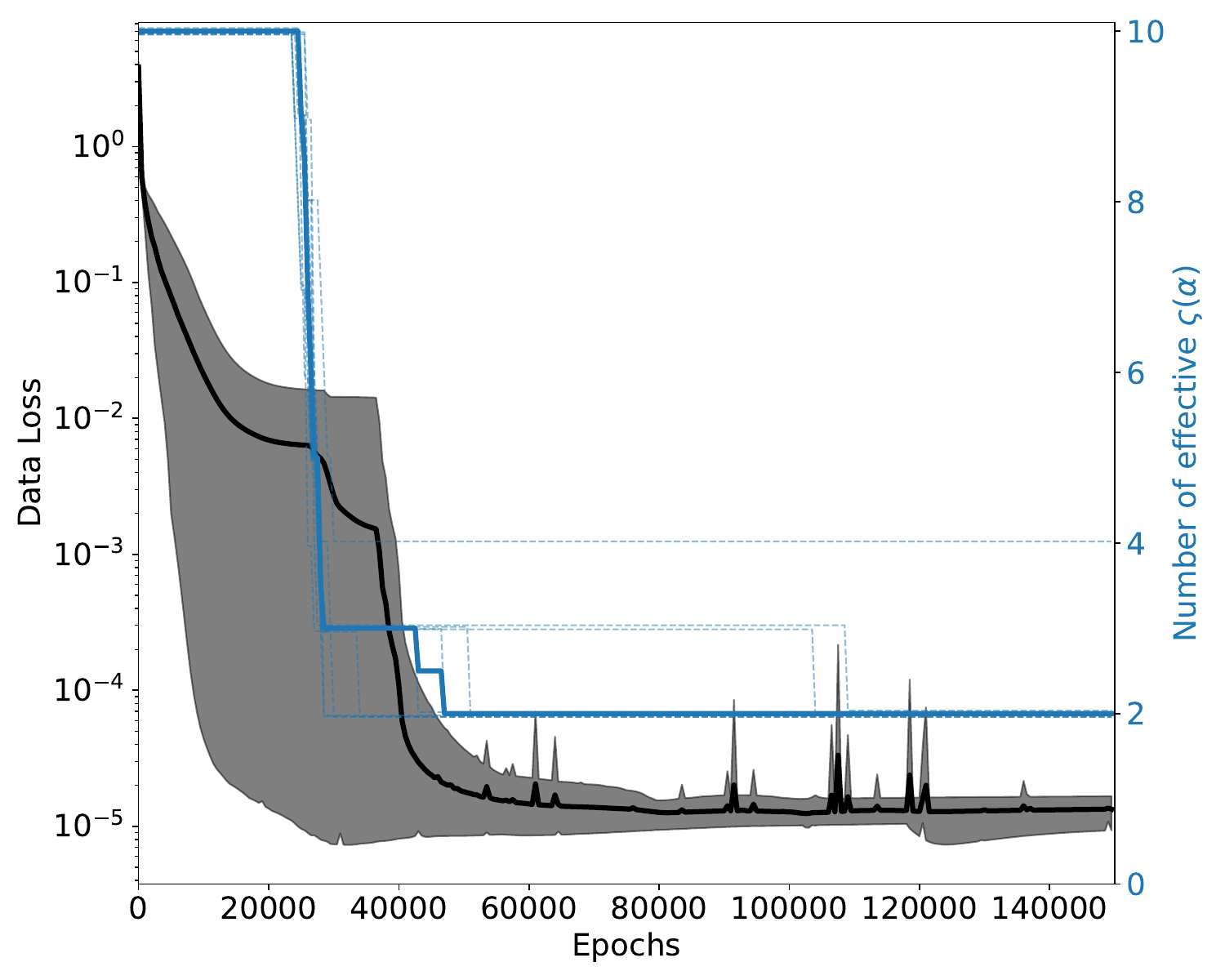}
\caption{double well}
\end{subfigure}
\begin{subfigure}[c]{0.32\textwidth}
\includegraphics[width=0.99\linewidth]{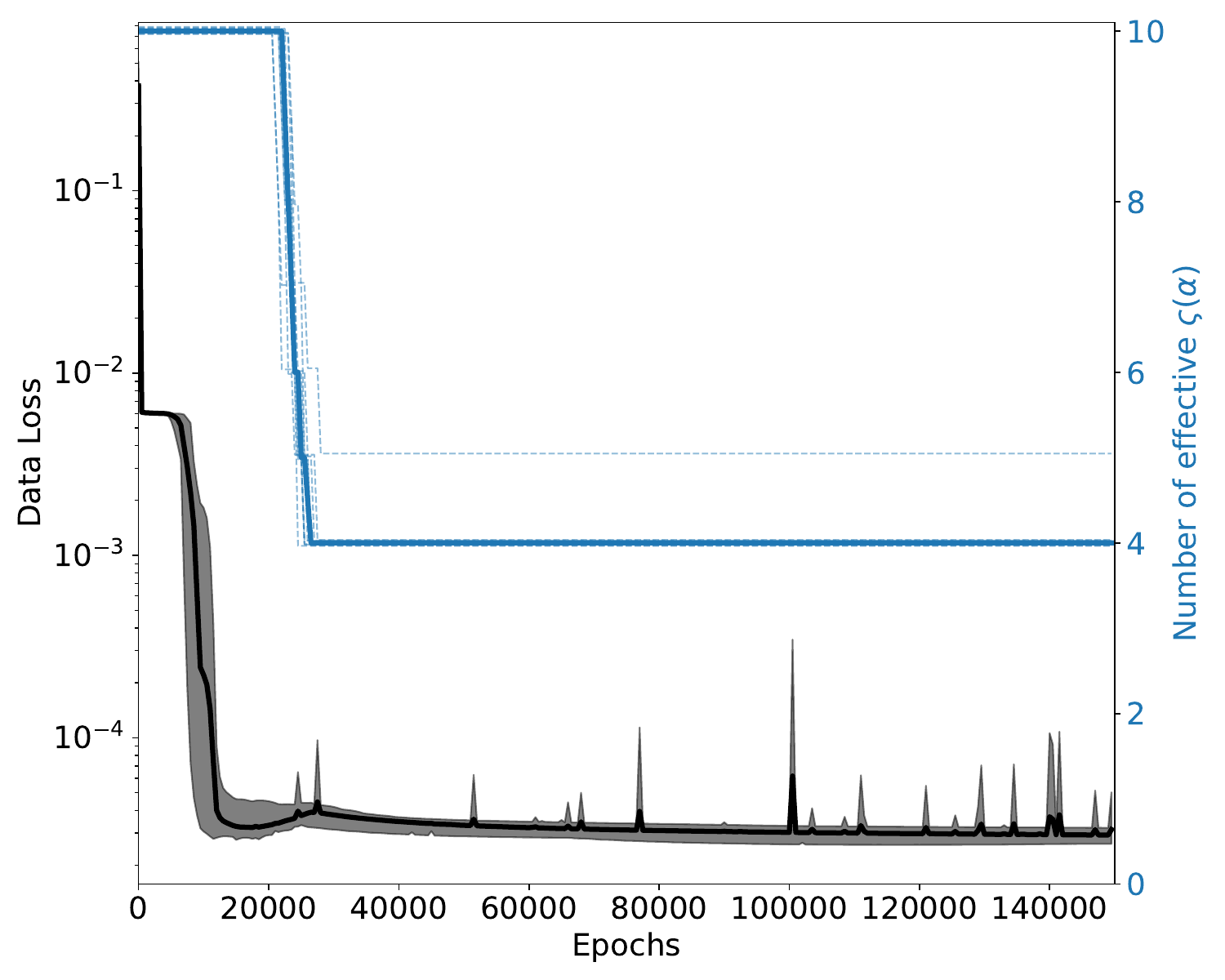}
\caption{modulated well}
\end{subfigure}
\begin{subfigure}[c]{0.32\textwidth}
\includegraphics[width=0.99\linewidth]{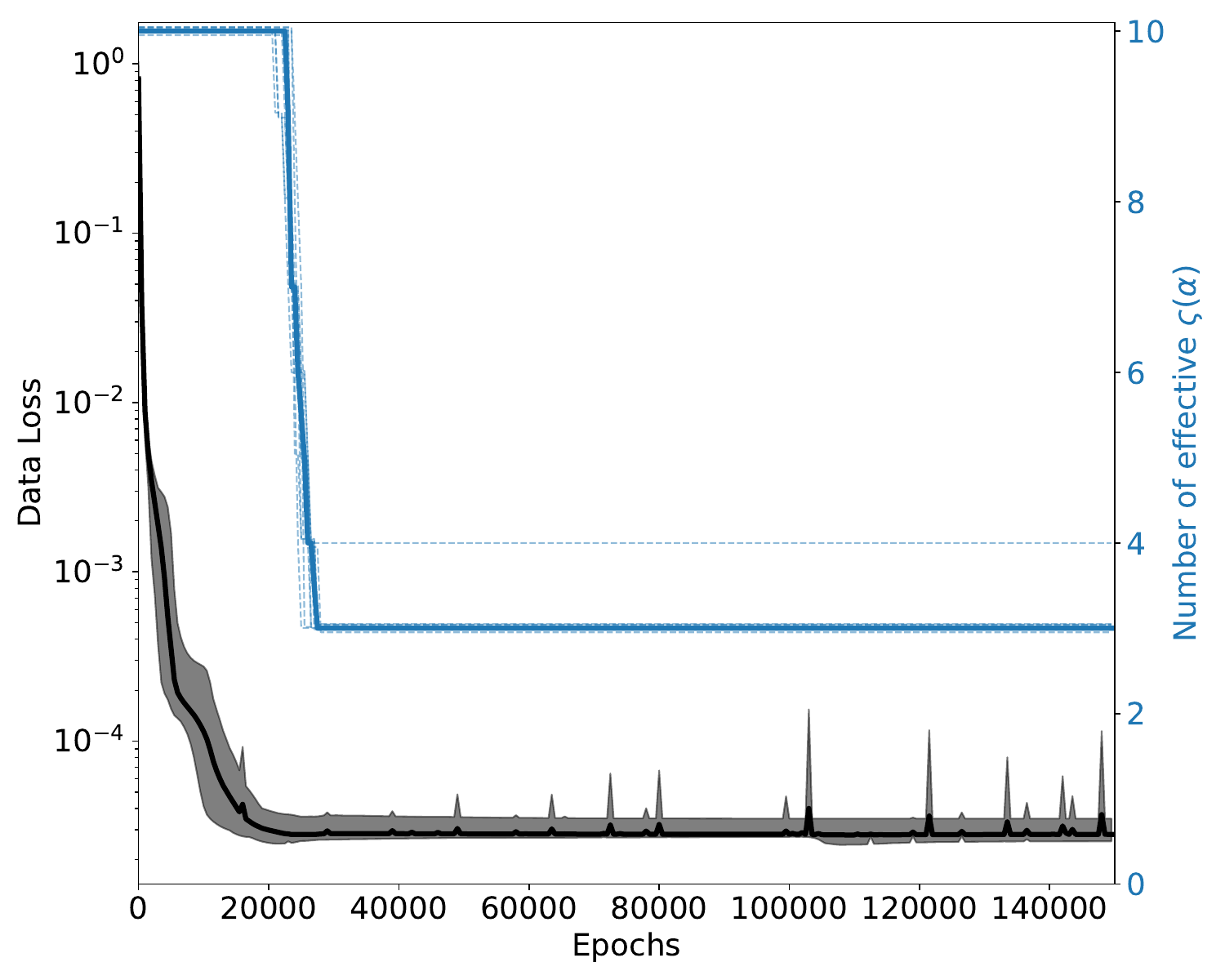}
\caption{minimum well}
\end{subfigure}
\caption{Multi-well potentials: error (black) and well count (blue) convergence for multiple calibrations. Median value is shown with a thicker line.}
\label{fig:1D_convergence}
\end{figure}

\begin{figure}[h]
\centering
\begin{subfigure}[c]{0.32\textwidth}
\includegraphics[width=0.99\linewidth]{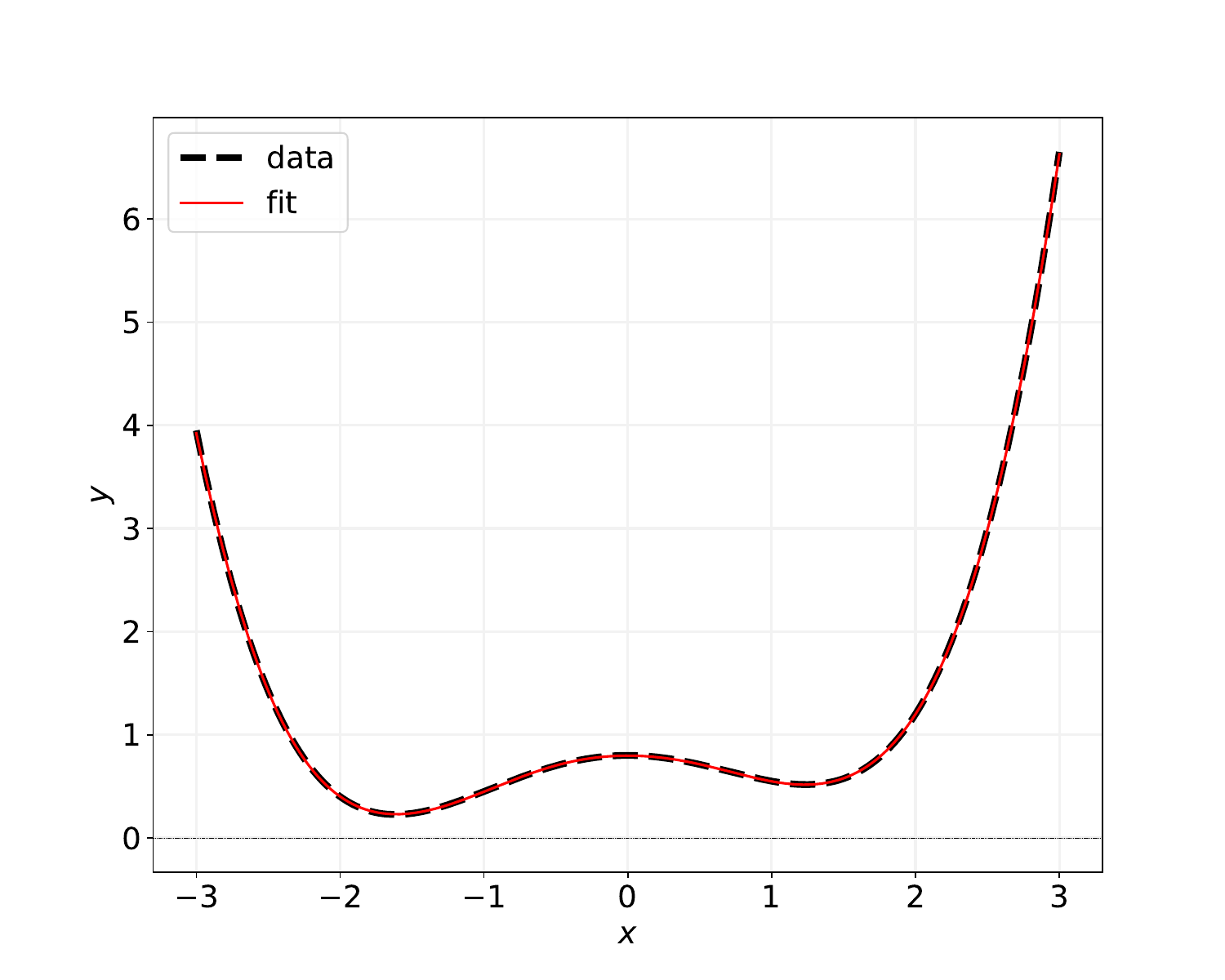}
\caption{double well}
\end{subfigure}
\begin{subfigure}[c]{0.32\textwidth}
\includegraphics[width=0.99\linewidth]{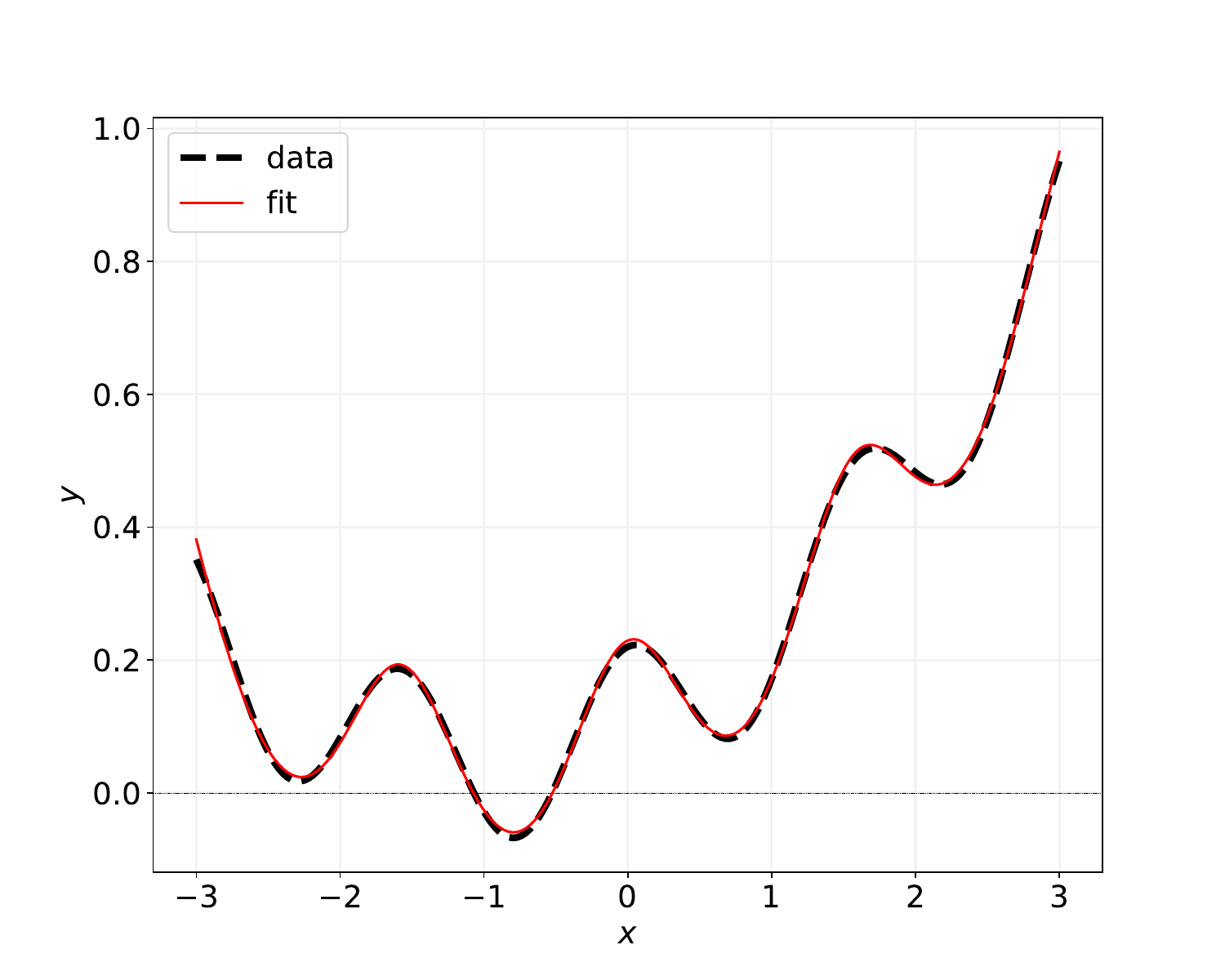}
\caption{modulated well}
\end{subfigure}
\begin{subfigure}[c]{0.32\textwidth}
\includegraphics[width=0.99\linewidth]{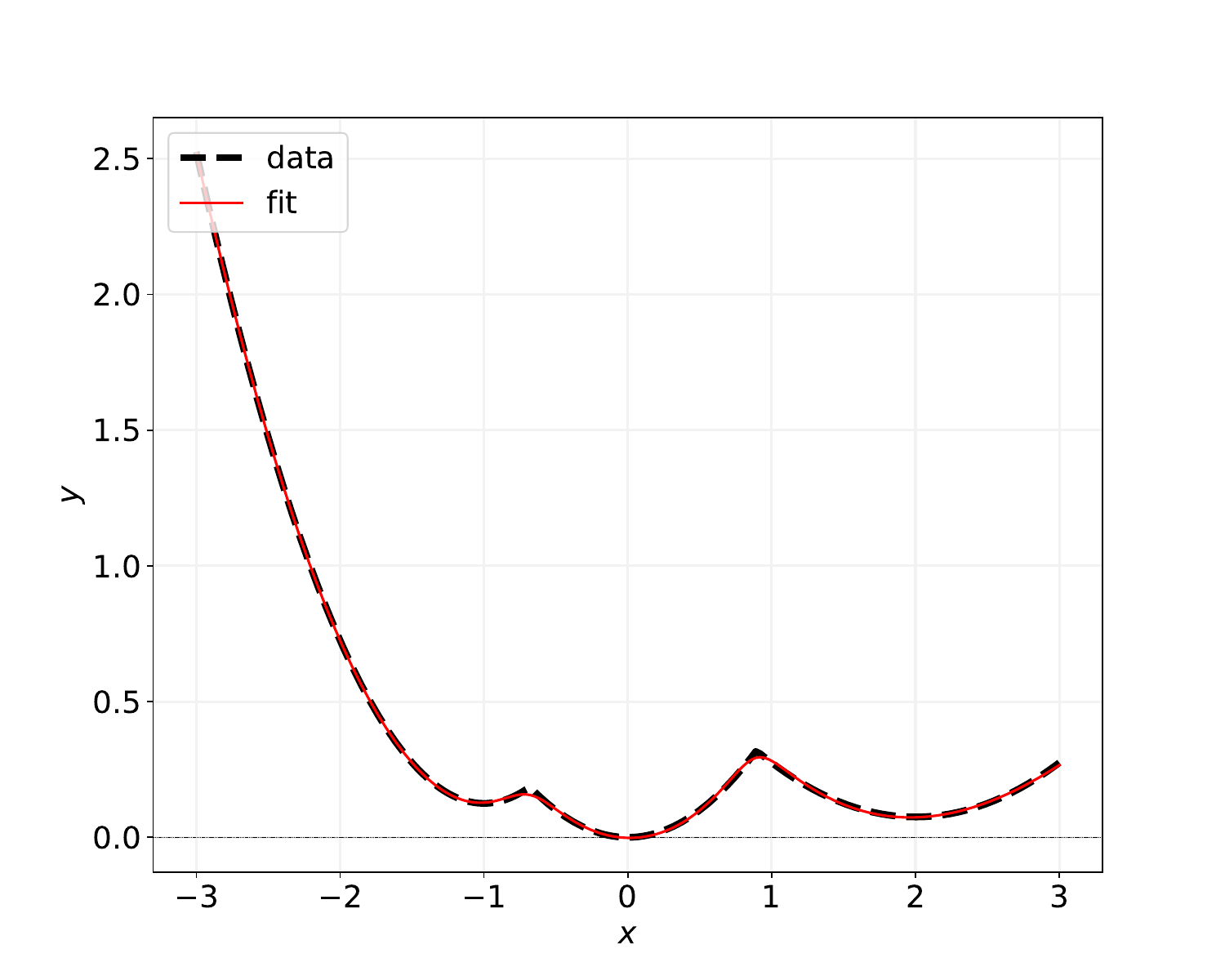}
\caption{minimum well}
\end{subfigure}
\caption{Multi-well potentials fits. Data: black dashed thick lines, fit: red thin lines.}
\label{fig:1D_demos}
\end{figure}

\begin{figure}[h]
\centering
\begin{subfigure}[c]{0.32\textwidth}
\includegraphics[width=0.99\linewidth]{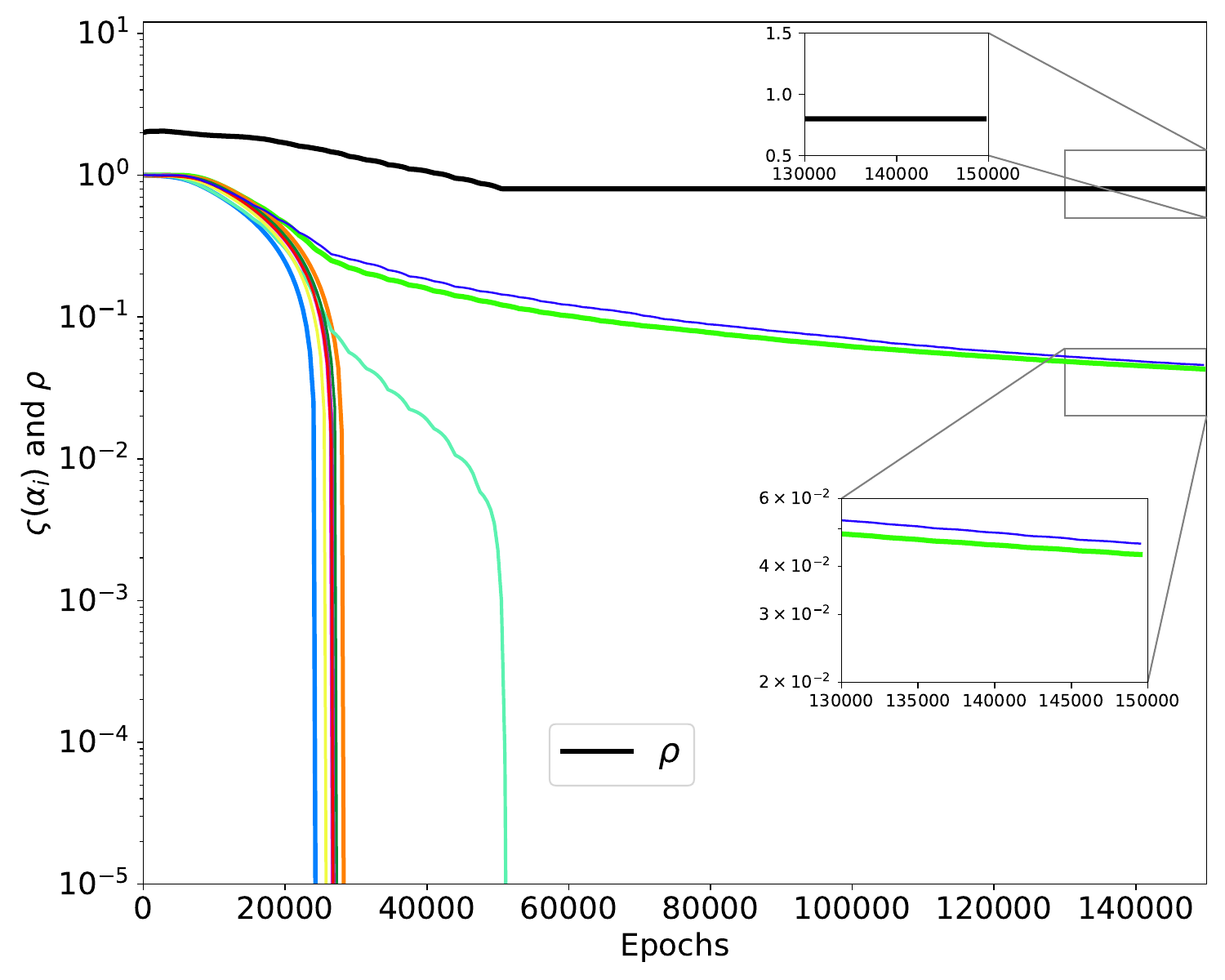}
\caption{double well}
\end{subfigure}
\begin{subfigure}[c]{0.32\textwidth}
\includegraphics[width=0.99\linewidth]{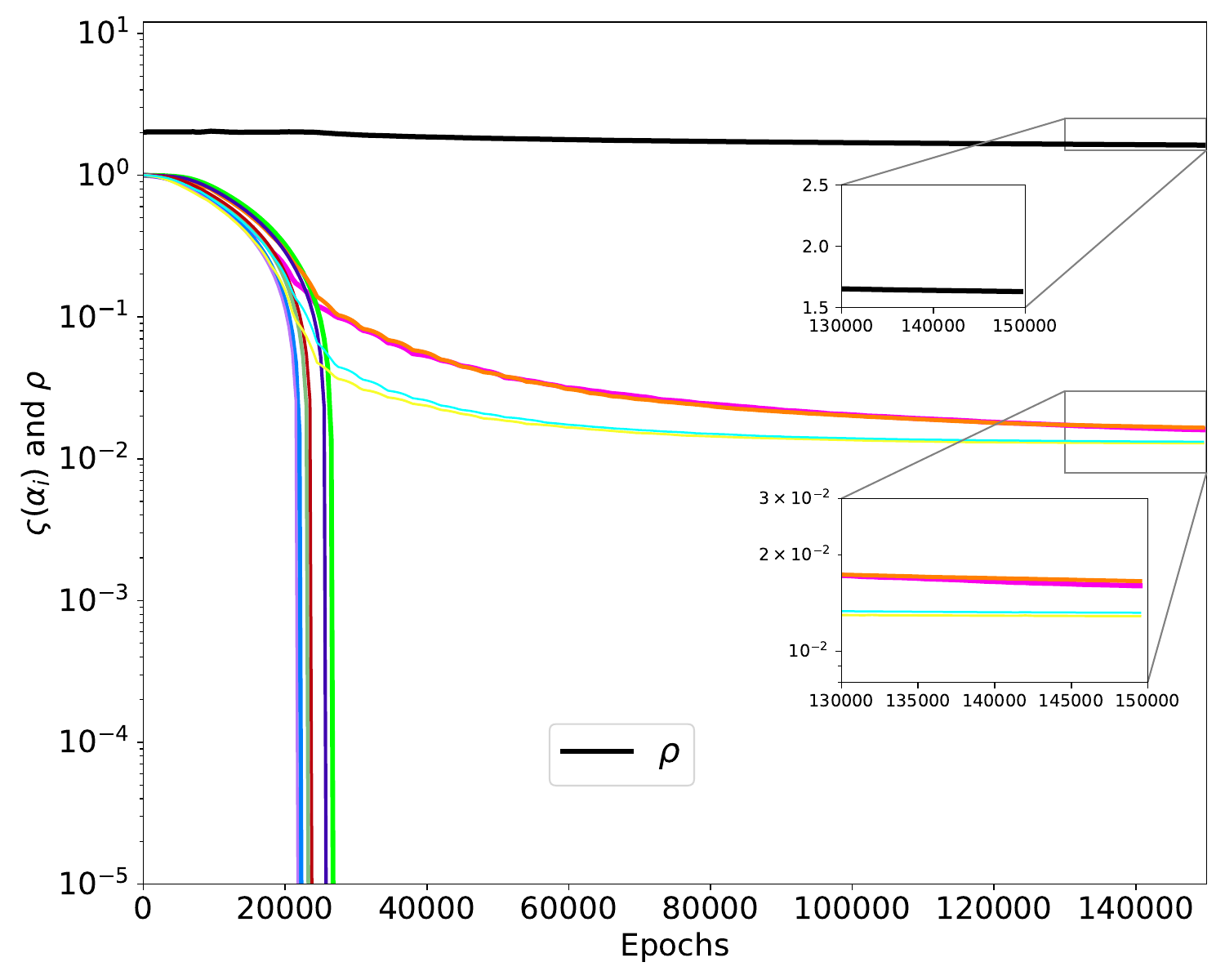}
\caption{modulated well}
\end{subfigure}
\begin{subfigure}[c]{0.32\textwidth}
\includegraphics[width=0.99\linewidth]{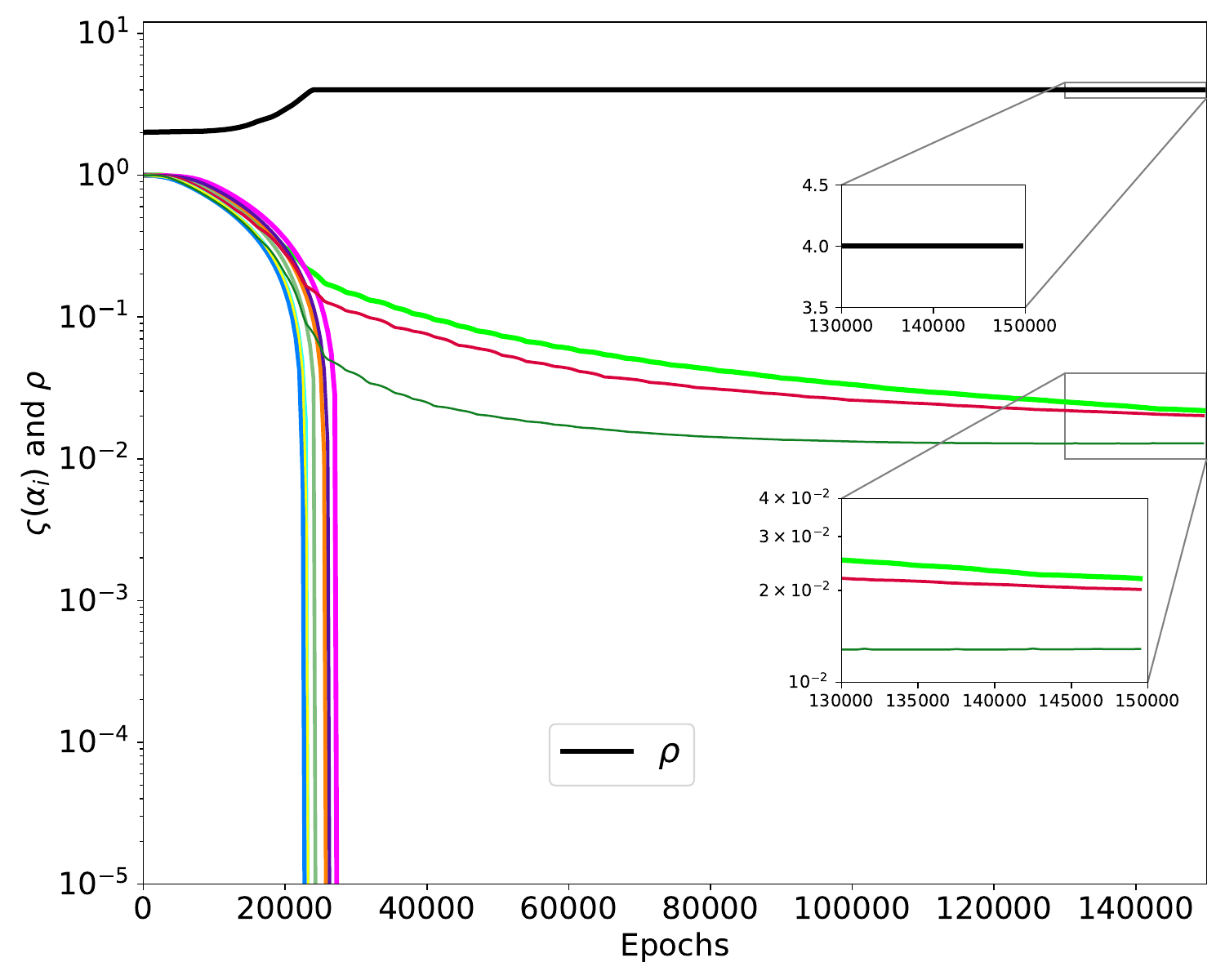}
\caption{minimum well}
\end{subfigure}
\caption{Multi-well potentials: evolution of scale $\rho$ and weights $\{ \alpha_i \}$ during training. Black curve depicts $\rho$; Multi-colored curves depicts the 10 $\{ \alpha_i \}$.}
\label{fig:1D_hist}
\end{figure}

For each, we used an LSE-ICNN with 2 layers with 10 nodes per layer and a maximum of 10 modes. We furthermore repeat the training process 10 times with different random seeds of the neural network weights and biases. We initially set $\varsigma(\alpha_i)=0.99$ for all $i=1,\ldots,10$ and set $\rho=2$. We employ the ADAM optimizer \cite{diederik2014adam} with a learning rate of $10^{-3}$ for the neural network parameters and $10^{-4}$ for the $\alpha_{i}$ and $\rho$.
We fix the regularization constant to $\epsilon=10^{-4}$. We train over 150k epochs.
\fref{fig:1D_convergence} shows the convergence of the effective number of active $\varsigma(\alpha_{i})$ (blue curves) for the 10 random initializations of the LSE-ICNN for each of the examples, as well as the loss (black curves). We define a parameter to be active if $\varsigma(\alpha_{i})>10^{-6}$. The thick lines indicate the median over the 10 runs.
It can be seen that the framework
generally identifies the correct number of modes and has fairly regular convergence. In particular, 1 out of 10 initializations ends up overestimating the number of wells for all three examples.
\fref{fig:1D_demos} plots the predicted and ground truth response of the three cases for a randomly picked initialization set. It can be seen that the LSE-ICNN representation not only captures the functions near the minima but also the relative smoothness of the local maxima, which are important in modeling the energetics of state transitions, for example.
Finally, the fairly rapid evolution of the scale and regularization parameters, $\rho$ and $\varsigma(\alpha_i)$, over the training process is shown in \fref{fig:1D_hist}.

\section{Demonstrations} \label{sec:results}

In this section, we demonstrate the representative power of the LSE-ICNN with demonstrations from mechanochemistry, material phase mechanics, biological evolution, and variational inference.

\subsection{Mechanochemical phase change}
A mechanochemical process has a free energy potential $\Psi$ dependent on deformation and chemical concentration ~\cite{garikipati2017perspectives,rudraraju2016mechanochemical,teichert2023bridging}.
Phase changes can occur when there are multiple minima in the free energy of the material.

\begin{figure}[h]
\centering
\includegraphics[width=0.85\linewidth]{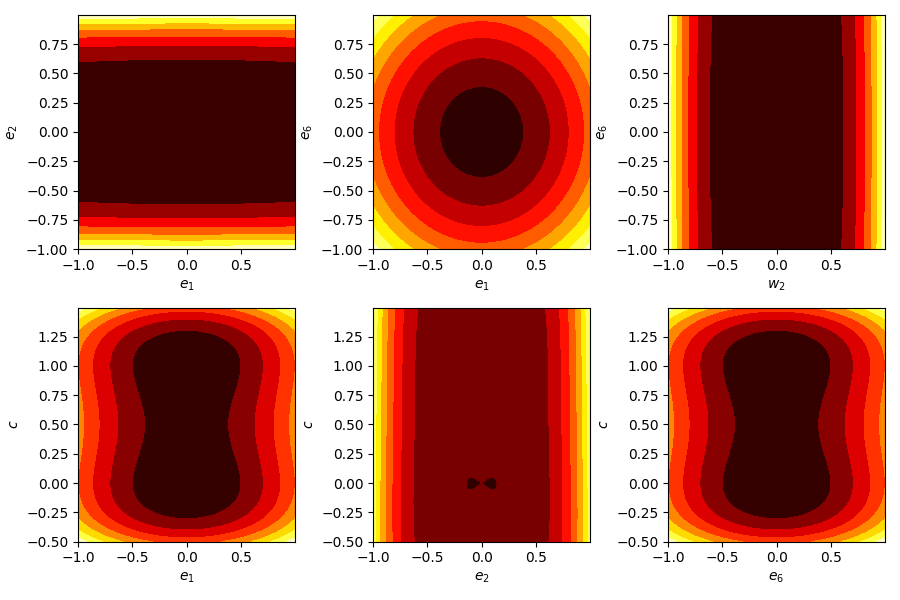}
\caption{Mechanochemical potential projections plotted via the invariants:
$e_1 = \frac{1}{\sqrt{3}} \tr \Eb$,
$e_2 = \frac{1}{\sqrt{2}}(E_{11}-E_{22})$,
$e_6 = \sqrt{2} E_{12}$,
and $c$.}
\label{fig:mechchem_potential}
\end{figure}

In this demonstration, we use the free energy from~\cref{rudraraju2016mechanochemical}:
\begin{eqnarray}
\Psi(\Eb,c) &=& d_c (96 c^2 - 192 c^3 + 96 c^4)  \\
&&+ 2 \sqrt{\frac{2}{3}}
\frac{d_e} {s_e^3} (c-1)
\left(
(  E_{11} +   E_{22} - 2 E_{33})
(  E_{11} - 2 E_{22} +   E_{33})
(2 E_{11} -   E_{22} -   E_{33}\right) \nonumber \\
&+& 6 \frac{d_e}{s_e^2} (2 c -1)
\left(E_{11}^2 + E_{22}^2 + E_{33}^2  - E_{22} E_{33} - E_{11} E_{22}  - E_{11} E_{33})\right)  \nonumber \\
&+& 4\frac{d_e}{s_e}
\left(E_{11}^2 + E_{22}^2 + E_{33}^2 - E_{22} E_{33} - E_{11} (E_{22} + E_{33})^2)\right) \nonumber \\
&+& \frac{d_e}{2 s_e}
\left(6 (E_{12}^2 + E_{13}^2 + E_{23}^2) + (E_{11} + E_{22} + E_{33}^2)\right) \nonumber \quad ,
\end{eqnarray}
where
$d_c = 2.0$, $d_e=0.1$, $s_e = 0.1$,
as a data-generating model.
Here,
$\Eb$ is the Lagrange strain and $c$ is a concentration.
The free energy has multiple wells, which are characteristic of a material that can undergo a phase change. 
The projections shown in \fref{fig:mechchem_potential} indicate that the mechanochemical potential $\Psi$ is highly nonlinear and globally non-convex with multiple wells.

We train an LSE-ICNN
\begin{equation}
\Psi = \LSE(\strain,c)
\end{equation}
to the observable derivatives: the stress tensor,
\begin{equation}
\stress  = \partialb_\strain \Psi ,
\end{equation}
and the chemical potential,
\begin{equation}
\mu = \partial_c \Psi \quad .
\end{equation}
For this demonstration, the LSE-ICNN had (maximum) $N_\text{modes}=$ 5 modes each with an ICNN with 6 strain $\Eb$ and 1 concentration $c$ input, 2 layers with 10 neurons each, and a single output.
The data $\Ds = \{ \Eb_i, c_i, \stress_i, \mu_i \}_i$, with inputs $\Xs_i = (\strain_i, c_i)$ and corresponding outputs $\Ys_i = \partialb_\Xs \Psi(\Xs_i) = (\stress_i, \mu_i)$ is split into 800 training and 200 testing samples.
Since we combine the (normalized) inputs and outputs, we employ an L2 loss $\Lc$ that equally weights the errors in the stress $\stress$ and the chemical potential $\mu$
\begin{eqnarray}
\MSE(\parameters; \data )
&=& \sum_i \left(\Ys_i - \partialb_\Xs \hat{\Psi}(\Xs_i; \parameters)\right)^2 \\
&=& \sum_i \left(\stress_i - \partialb_\strain \hat{\Psi}(\strain_i,c_i; \parameters)\right)^2
+ \sum_i \left(\mu_i - \partial_c \hat{\Psi}(\strain_i,c_i; \parameters)\right)^2 \quad .
\nonumber
\end{eqnarray}

\begin{figure}[h]
\centering
\begin{subfigure}[c]{0.48\linewidth}
\includegraphics[width=0.85\linewidth]{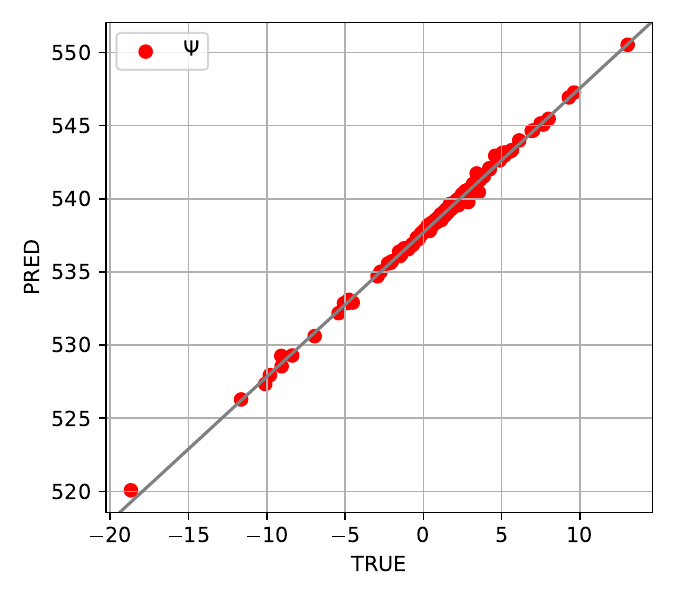}
\caption{potential}
\end{subfigure}
\begin{subfigure}[c]{0.48\linewidth}
\includegraphics[width=0.85\linewidth]{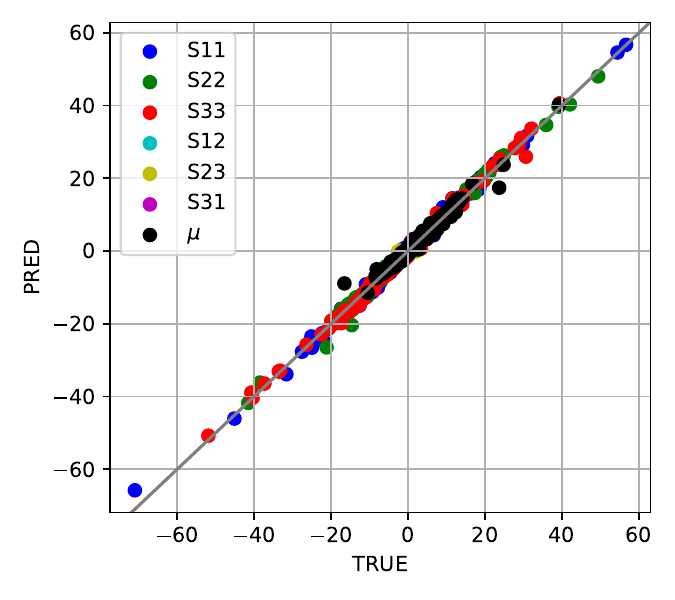}
\caption{gradient}
\end{subfigure}
\caption{Mechanochemical correlations with held-out test data. Note that the offset in the predicted vs. true potential does not affect the potential gradients since the true and predicted potentials are highly correlated.}
\label{fig:mechchem_test}
\end{figure}

\begin{figure}[h]
\centering
\begin{subfigure}[c]{0.48\linewidth}
\includegraphics[width=0.85\linewidth]{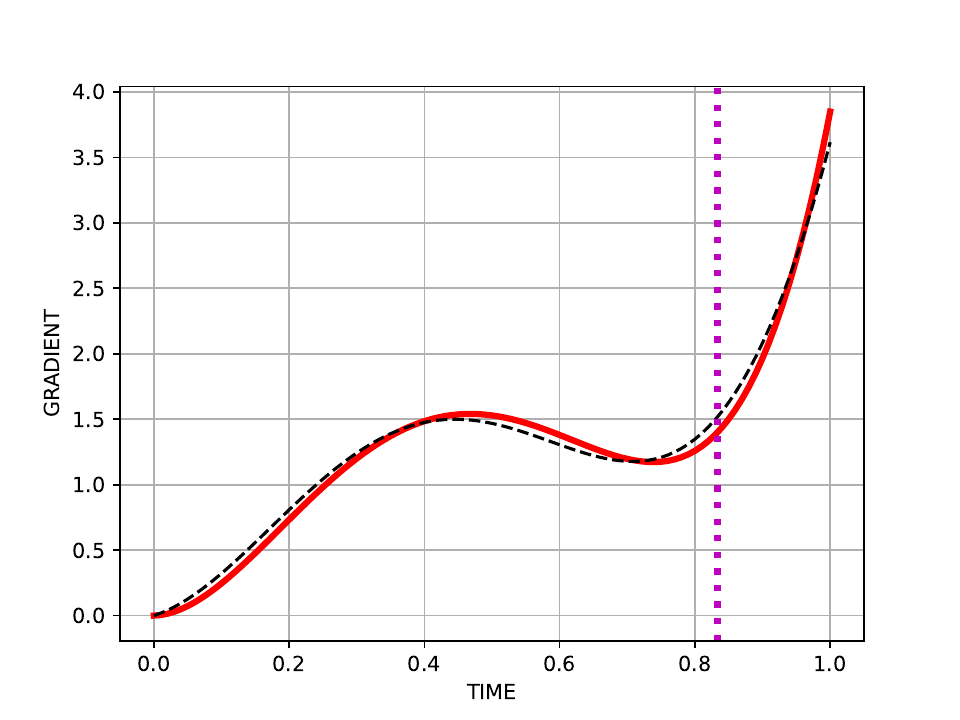}
\caption{potential}
\end{subfigure}
\begin{subfigure}[c]{0.48\linewidth}
\includegraphics[width=0.85\linewidth]{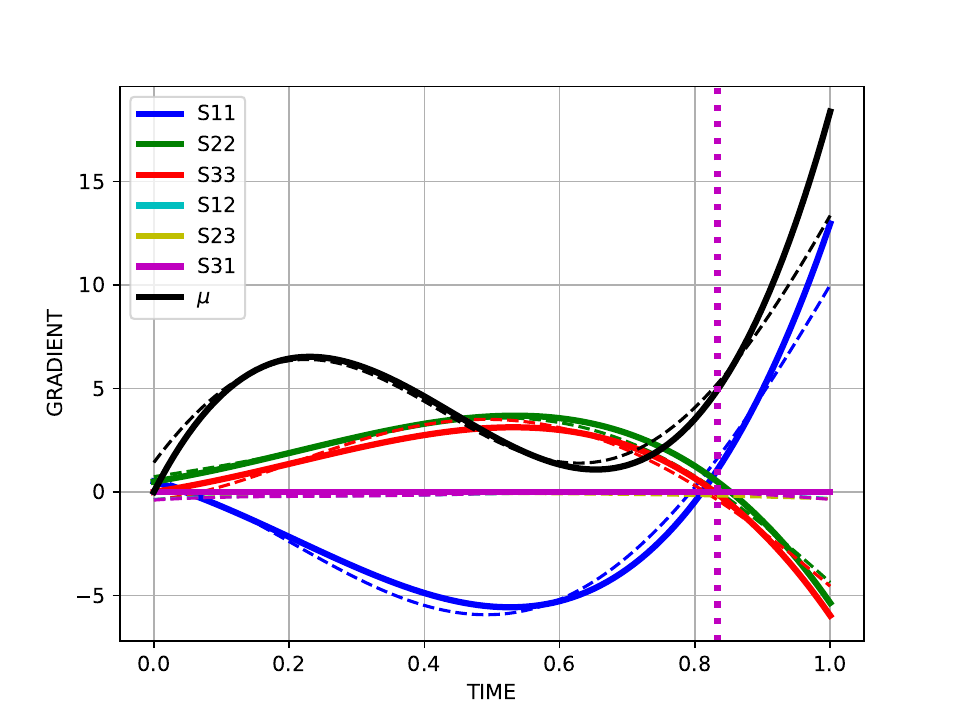}
\caption{gradient}
\end{subfigure}
\caption{Mechanochemical comparison of predictions (dashed lines) with validation data (solid lines). The model is in extrapolation past the vertical dotted line.}
\label{fig:mechchem_validation}
\end{figure}

The parity plots in \fref{fig:mechchem_test}(a,b) show that the LSE-ICNN provides an excellent prediction of the inferred potential and gradient used in training on held-out data.
Furthermore, \fref{fig:mechchem_validation} illustrates that the ICNN-based LSE has good extrapolation properties for both the gradient and inferred potential on an unseen strain-concentration trajectory:
\begin{equation}
(\strain(t),c(t)) =  \Bigl( 0.4 \, t \bigl( \eb_1\otimes\eb_1-0.3(\Ib-\eb_1\otimes\eb_1)\bigr), t
\Bigr)
\quad \text{where} \quad t \in [0,1] \ .
\end{equation}

\subsection{Multimodal distributions}

Representing the probability distribution of stochastic bifurcating dynamics in chemical and other physical systems represents a challenge.
The Schl\"{o}gl system involves three species and two reactions and is a canonical chemical reaction that exhibits bistability~\cite{schlogl1972chemical,vellela2009stochastic}.

The Schl\"{o}gl model is a prototypical bistable chemical reaction system involving a single dynamic species \( X \) and two reservoir species \( A \) and \( B \), which are maintained at constant concentrations. The model consists of the following reactions:

\begin{align}
A + 2X & \underset{k_2}{\overset{k_1}{\rightleftharpoons}} 3X \label{eq:schlogl-reaction1} \\
B & \underset{k_4}{\overset{k_3}{\rightleftharpoons}} X \label{eq:schlogl-reaction2}
\end{align}

Assuming constant concentrations \( [A] = a \) and \( [B] = b \), the concentration \( x(t) = [X](t) \) evolves according to the ordinary differential equation:
\begin{equation}
\frac{dx}{dt} = k_1 a x^2 - k_2 x^3 + k_3 b - k_4 x \quad .
\label{eq:schlogl}
\end{equation}

\begin{figure}[h]
\centering
\begin{subfigure}[c]{0.32\textwidth}
\includegraphics[width=0.99\linewidth]{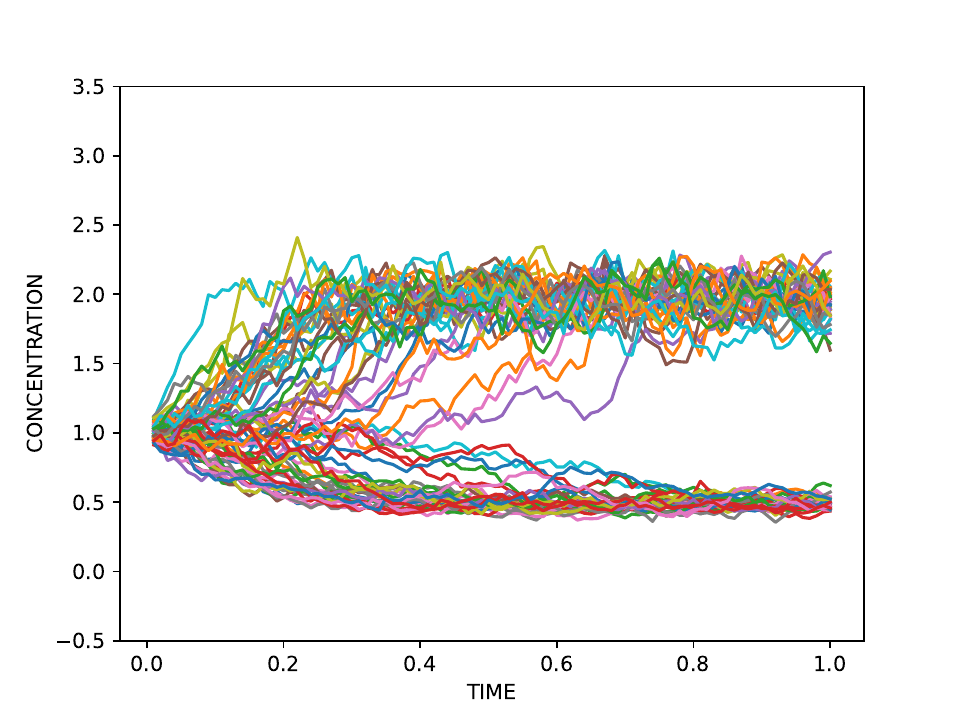}
\caption{trajectories}
\end{subfigure}
\begin{subfigure}[c]{0.32\textwidth}
\includegraphics[width=0.99\linewidth]{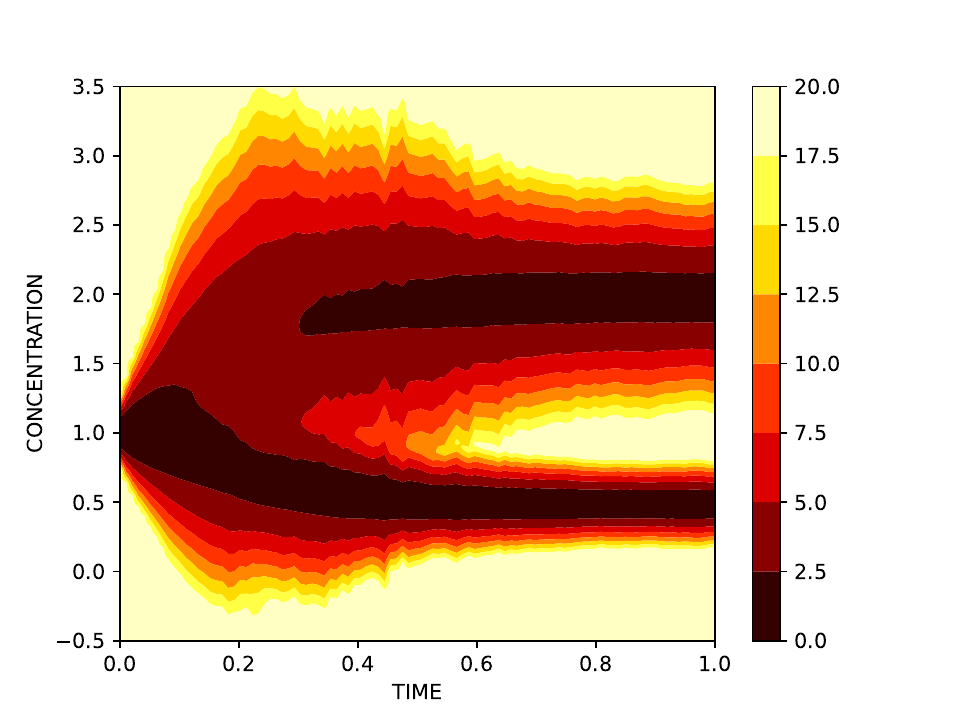}
\caption{data PDF}
\end{subfigure}
\begin{subfigure}[c]{0.32\textwidth}
\includegraphics[width=0.99\linewidth]{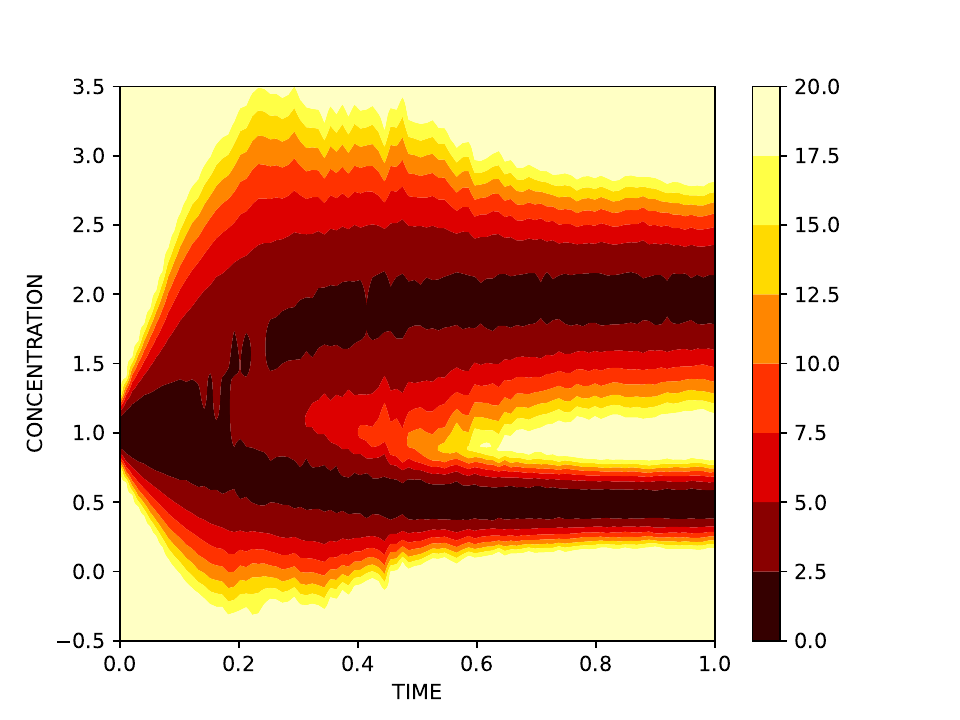}
\caption{fitted PDF}
\end{subfigure}
\caption{Schlogl reaction: (a) data, (b) KDE of data, (c) LSE fit of data.}
\label{fig:schlogl}
\end{figure}

In the data-generating model, the reaction kinetics transition probabilities drive a Gillespie stochastic simulation algorithm \cite{gillespie1977exact,gillespie2007stochastic}.
The output of the SSA for the Schl\"{o}gl system is multimodal and depends on the particular realization of the stochastic diffusion process and the initial conditions.
More details can be found in \cite{sargsyan2010spectral,safta2025uncertainty}.
\fref{fig:schlogl}a shows a random selection of the 200 trajectories used to fit an LSE-ICNN representation.
The higher concentration mode is broader/more diffuse than the lower concentration mode, which is more apparent in the kernel density estimate (KDE) shown in \fref{fig:schlogl}b.

To model this data, we use variational inference (VI) \cite{blei2017variational} with an LSE-ICNN representation of the probability density function (PDF) $\prob(y)$ of $y$, the single observed species:
\begin{equation}
\log \prob(y) \approx \log q(y; \parameters) \equiv \text{LSE}(y; \parameters) \quad .
\end{equation}
In this learning framework $q$ plays the role of the posterior density $\prob(\parameters \ | \ y)$ given data $y$ as in Bayes rule:
\begin{equation}
\underbrace{ \prob(\parameters \ | \ y) }_\text{posterior} \propto
\underbrace{ \prob(y \ | \ \parameters)  }_\text{likelihood}
\underbrace{ \prob(\parameters)  }_\text{prior}
\end{equation}
Minimizing the (negative) evidence lower bound (ELBO) provides an optimization problem
\begin{equation} \label{eq:elbo}
\parameters(t) = \argmin_\parameters \Bigl[ -\text{ELBO}(\parameters ; y)  \Bigr]
= \argmin_\parameters \Bigl[  -\Ebb_{q} \left[ \log \prob(y(t) \ | \ \parameters) \right]
+ \operatorname{D}_\text{KL} (q(y; \parameters) || \prob(y)) \Bigr]
\end{equation}
for calibrating the parameters of the LSE-ICNN posterior distribution.
In \eref{eq:elbo}, the first term is the negative log-likelihood, which is proportional to the MSE and therefore drives data fit, while the second term is the Kullback-Leibler divergence between the variational posterior $q(y; \parameters)$ and the prior $\prob(y)$, which we assume is uniform and zero-valued.
The KL divergence term provides a regularization of the main MSE-like objective.

For this application, we used a (maximum) $N_\text{mode}=$ 5 mode LSE-ICNN with a single input per time step (the observed species concentration $y(t)$), one output (the probability density $q$),  two hidden layers with 10 neurons each.

The quantitative similarity between \fref{fig:schlogl}(b,c) demonstrates the success of the LSE representation (which is fitted per time-step) in this application.

\subsection{Microstructural elastic instability}

To demonstrate that an LSE-ICNN potential is a good representation of phase change and related material stability systems, we apply it to model the behavior of a metamaterial that experiences microstructural instabilities resembling classical phase change.
Rossi \etal \cite{rossi2024limit} provides a phenomenological model of the mechanical response of a metamaterial tuned to detailed microstructural simulations,  which we adapt to be fully 3D.
The (two-well) free energy is a function of strain $\strain$ and internal variable $c$, which determines the phase:
\begin{equation}
\Psi(\strain,c) = (1-c) \psi_A(\strain_A) + c \psi_B (\strain_B) \quad ,
\end{equation}
with $\psi_A$ and $\psi_B$ being the energies of the two phases.
Additionally, the mixture rule
$\strain = (1-c) \strain_A + c \, \strain_B$ and an equilibrium condition
\begin{equation}
\partialb_\strain \Psi_A(\strain_A) =
\partialb_\strain \Psi_B(\strain_B)
\end{equation}
hold.
The stress is given by
\begin{equation}
\stress = \partialb_\strain \Psi
= \Cbb (\strain - c \strain_0) \, ,
\end{equation}
where $\Cbb$ is the elastic modulus tensor for both phases and $\strain_0$ is a transformation strain,
and a chemical potential-like conjugate force
\begin{equation}
\kappa = -\partial_c \Psi
= \stress \cdot \strain_0 - \Delta\psi \ ,
\end{equation}
where $\Delta\psi$ is the difference in the free energy between phases B and A.
The two primary equilibria are at the pure phases:
(a) $c=0$ and $\strain = \mathbf{0}$, and
(b) $c=1$ and $\strain = \strain_0$.

The dissipation potential $\phi$ governing the behavior of $c$ is not smooth due to the constraint $c \in [0,1]$
\begin{equation}
\phi(c,\dot{c})
=
\begin{cases}
+k_{+}\, \dot{c}, & \text{if}  \, \, \dot{c} \ge 0 \ \text{and} \ c \in [0,1),   \\
-k_{-}\,  \dot{c}, & \text{if}  \, \,  \dot{c} < 0 \ \text{and} \ c \in (0,1],  \\
\infty, & \text{else}.
\end{cases}
\end{equation}
This results in the flow rule
\begin{equation}
\dot{c}
=
\begin{cases}
+k_{+}, &  \text{if} \, \,   \kappa = +k_+,   \\
-k_{-}, & \text{if} \, \,   \kappa = -k_- ,   \\
0, &\text{else} .
\end{cases}
\end{equation}

\tref{tab:stability_properties} provides the model parameters we employed.
Further details are given in \aref{app:elastic_stability}.

\begin{figure}
\centering
\begin{subfigure}[c]{0.32\textwidth}
\centering
\includegraphics[width=0.95\linewidth]{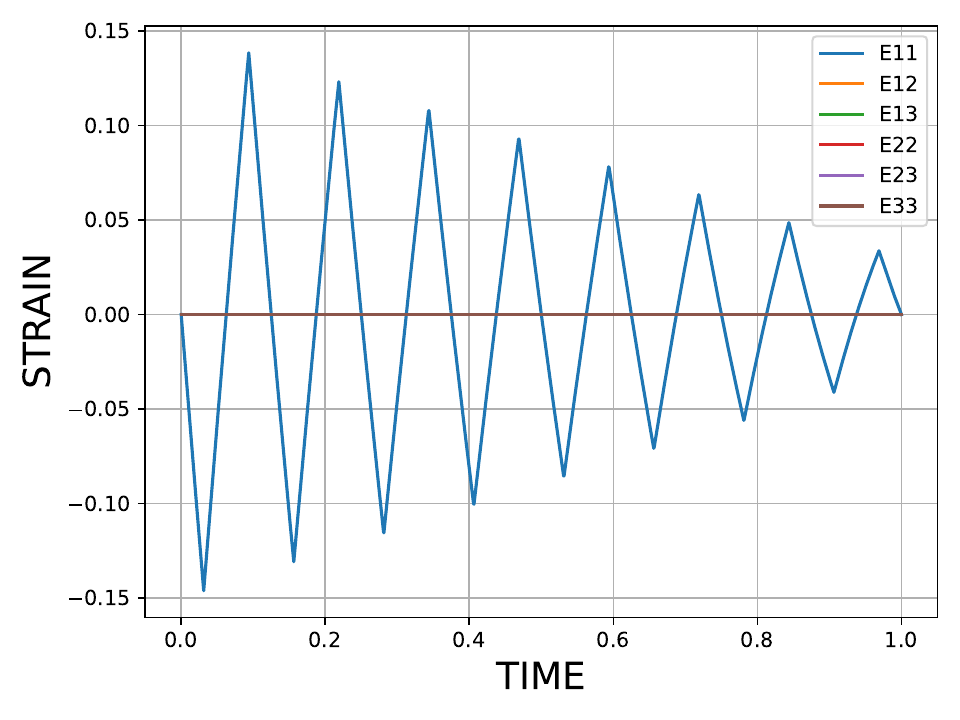}
\caption{strain-time}
\end{subfigure}
\begin{subfigure}[c]{0.32\textwidth}
\centering
\includegraphics[width=0.95\linewidth]{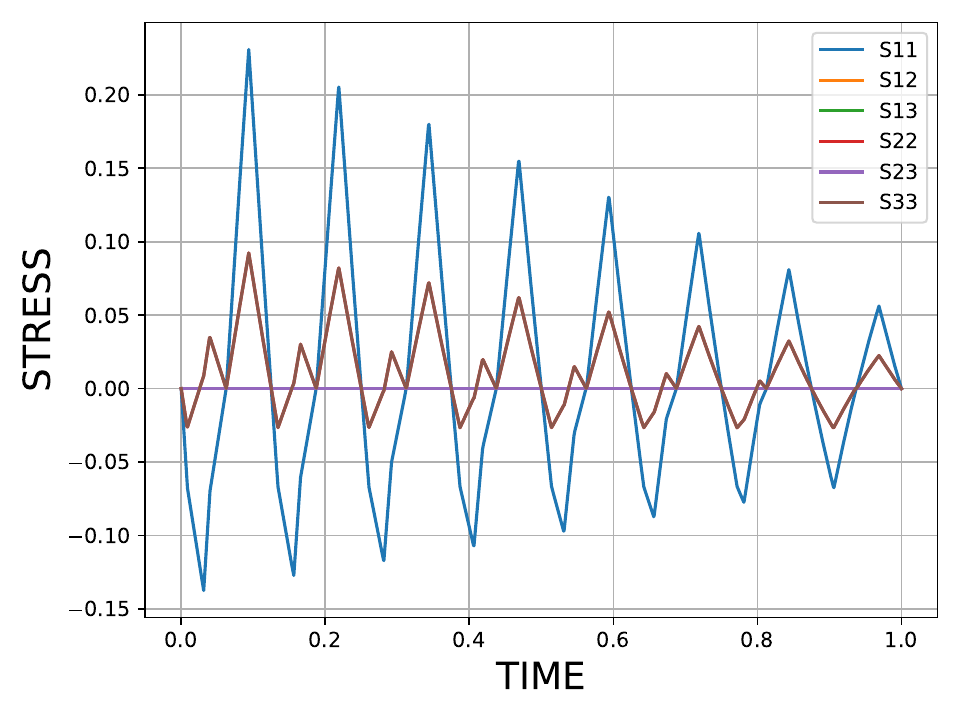}
\caption{stress-time}
\end{subfigure}
\begin{subfigure}[c]{0.32\textwidth}
\centering
\includegraphics[width=0.95\linewidth]{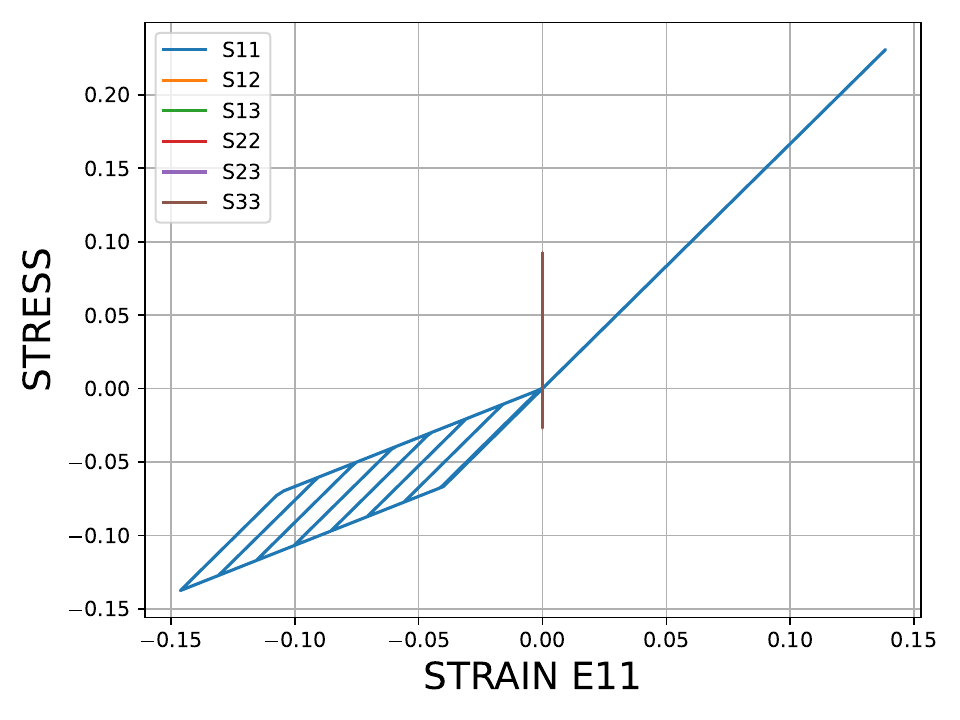}
\caption{stress-strain}
\end{subfigure}
\begin{subfigure}[c]{0.32\textwidth}
\centering
\includegraphics[width=0.95\linewidth]{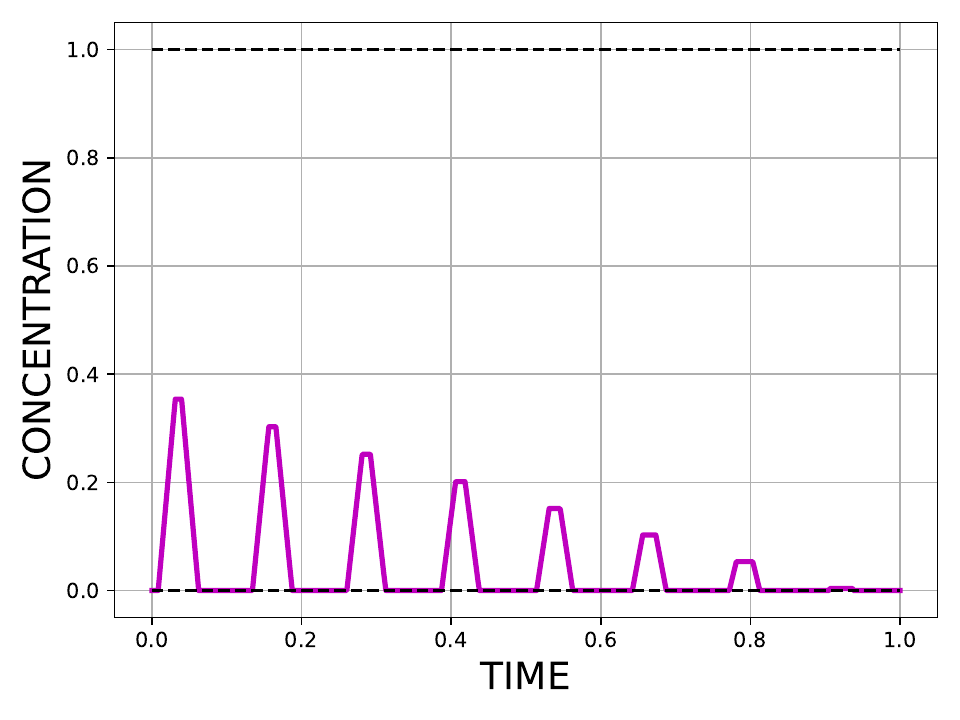}
\caption{concentration-time}
\end{subfigure}
\begin{subfigure}[c]{0.32\textwidth}
\centering
\includegraphics[width=0.95\linewidth]{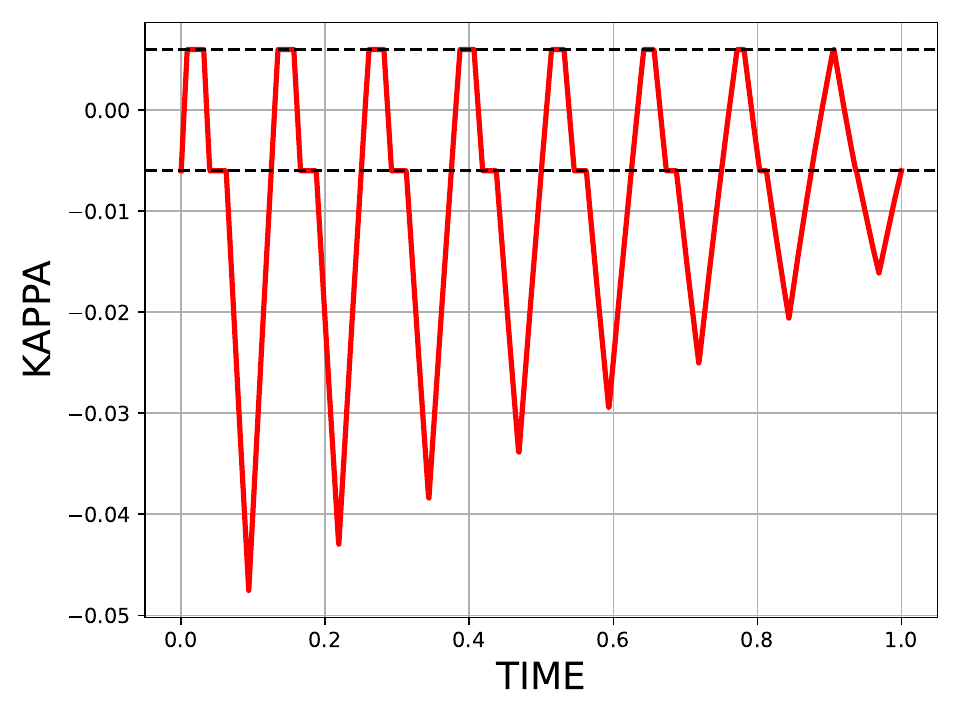}
\caption{conjugate force-time}
\end{subfigure}
\begin{subfigure}[c]{0.32\textwidth}
\centering
\includegraphics[width=0.95\linewidth]{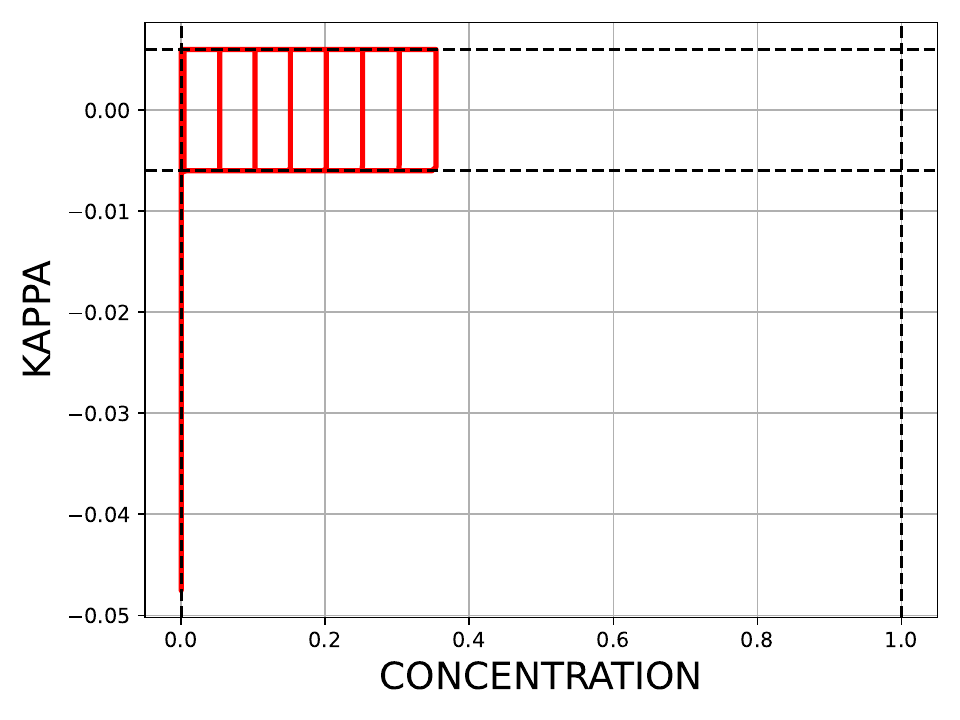}```
\caption{conjugate force-concentration}
\end{subfigure}
\caption{Elastic stability: representative data. Note only $E_{11}$ is non-zero.}
\label{fig:stability_data}
\end{figure}

We employed cyclic uniaxial sawtooth loading with randomized periods and amplitudes to generate train and test data mimicking experimental tests.
We split 400 loading trajectories into 80\% training and 20\% testing.
\fref{fig:stability_data} illustrates the model response to cyclic loading.
Notice the hysteresis cycles in the stress response due to the phase change facilitated by the internal variable $c$.

\begin{table}[]
\centering
\begin{tabular}{|c|c|c|c|c|}
\hline
$K_a$ & $G_a$ & $\epsilon_0 $ & $\Delta \psi$  & $k_\pm$   \\ \hline
1.0 & 0.5 &
-0.1 &
0.006 &
0.006 \\
\hline
\end{tabular}
\caption{Elastic stability: non-dimensionalized parameters}
\label{tab:stability_properties}
\end{table}

We modeled this data with an LSE-ICNN potential:
\begin{equation}
\Psi = \LSE(\inputvector,\hiddenvector) \, ,
\end{equation}
augmented by a neural ordinary differential equation (NODE) in the style of the ISV-NODE \cite{jones2021neural}:
\begin{equation}
\dot{\hiddenvector} = \NN(\inputvector,\hiddenvector) \, ,
\end{equation}
to provide the internal dynamics of phase change.
The NN  output is given by
\begin{equation}
\outputvector= \partialb_\inputvector \Psi(\inputvector,\hiddenvector) \, ,
\end{equation}
where $\hiddenvector$ is the hidden/internal state.
We used 2 modes in the LSE-ICNN potential because we know the number of phases and to limit the number of parameters.
Each LSE-ICNN mode had 3 inputs from the strain invariants plus 6 from the assumed internal state $\hiddenvector$, 2 hidden layers 19 neurons wide, and a single output: $\Psi$.
The NODE right-hand side network had 9 inputs from the strain and rate invariants, plus 6 from the assumed internal state, 2 hidden layers 15 neurons wide, and 6 outputs giving $\dot{\hiddenvector}$.
We assumed that the phase concentration $c$ was not observable and constructed an MSE on the stress $\stress$ alone using the strain $\strain$ as the input.

\begin{figure}
\centering
\begin{subfigure}[c]{0.45\linewidth}
\includegraphics[width=0.99\linewidth]{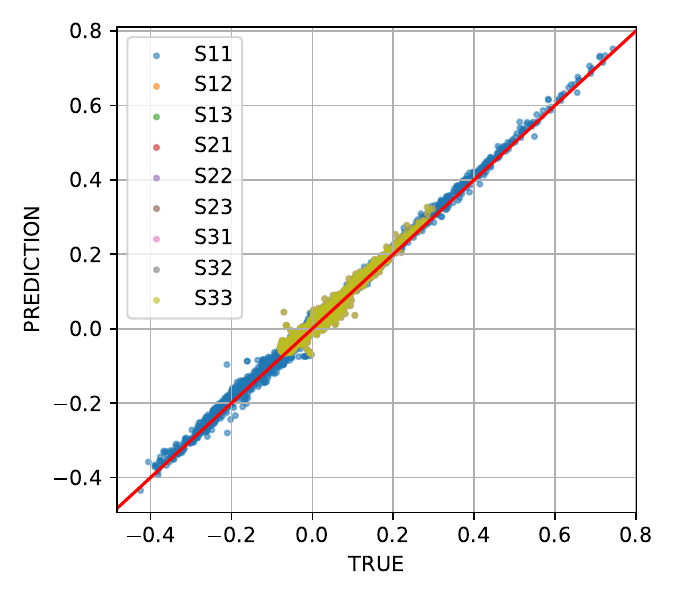}
\caption{fixed phase}
\end{subfigure}
\begin{subfigure}[c]{0.45\linewidth}
\includegraphics[width=0.99\linewidth]{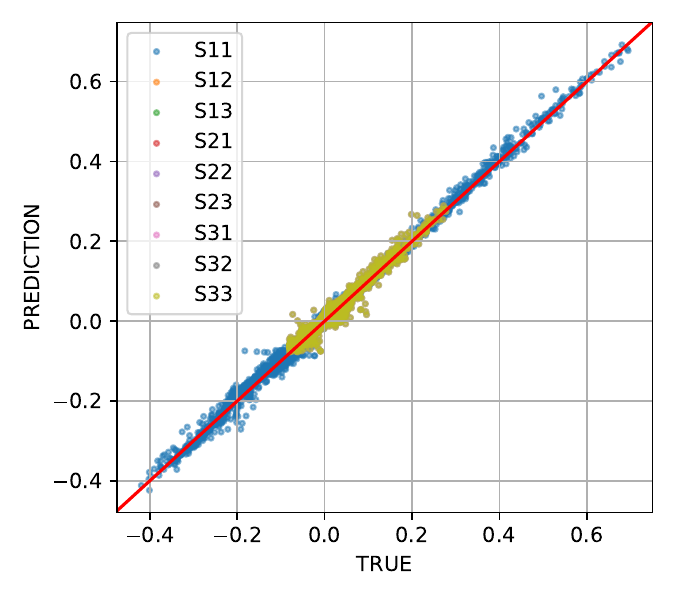}
\caption{2 phases}
\end{subfigure}
\caption{Elastic stability: parity between held-out data and model predictions.}
\label{fig:stability_correlation}
\end{figure}

\begin{figure}
\centering
\begin{subfigure}[c]{0.45\linewidth}
\includegraphics[width=0.99\linewidth]{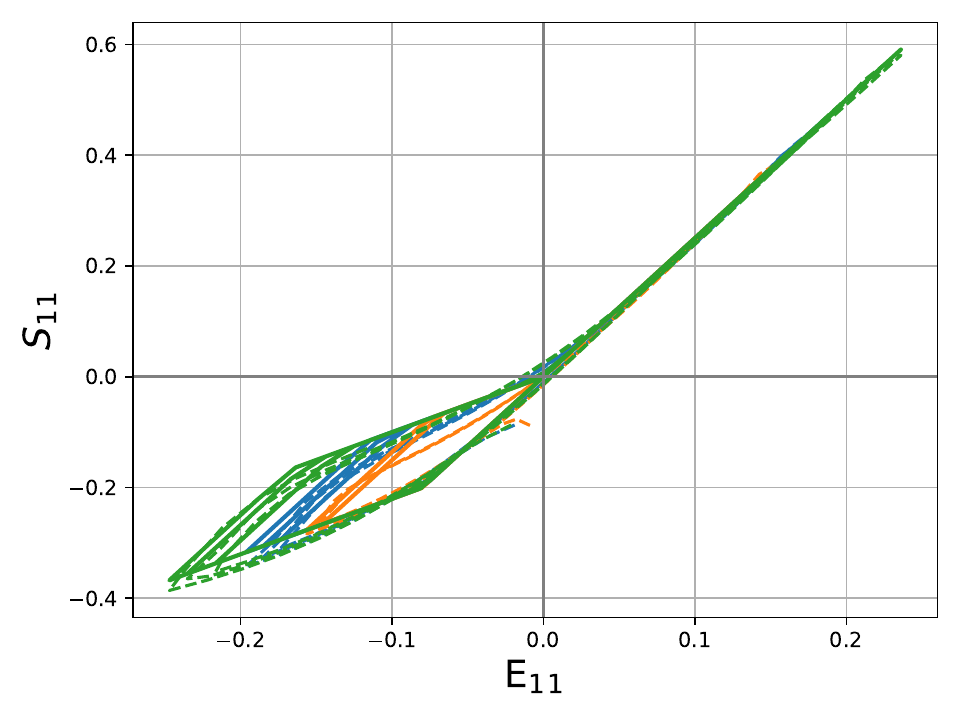}
\caption{}
\end{subfigure}
\begin{subfigure}[c]{0.45\linewidth}
\includegraphics[width=0.99\linewidth]{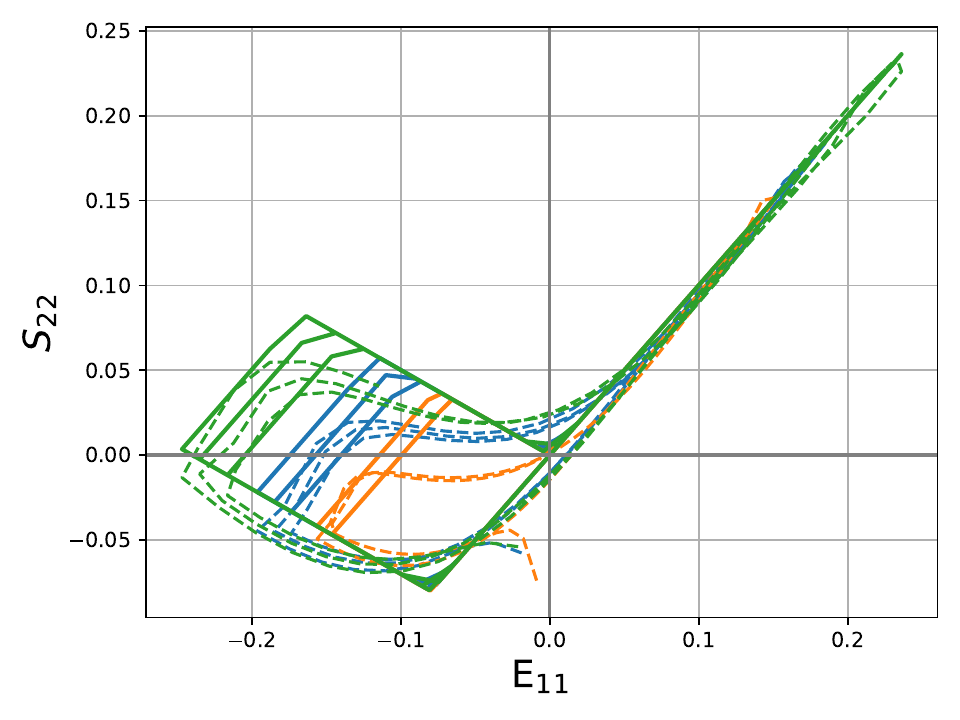}
\caption{}
\end{subfigure}
\caption{Elastic stability: fit (dashed lines) to held-out data (solid lines) for 3 trajectories}
\label{fig:stability_fit}
\end{figure}

\fref{fig:stability_correlation} shows the parity between true and predicted values on held-out test sets for two cases: (a) a fixed-phase material ($c=0.5$) and (b) the two-phase material with full internal dynamics.
Given the non-smooth data and the inherent smoothness of the NODE, the accuracy is good for both cases.
Finally, \fref{fig:stability_fit} compares held-out and predicted trajectories for the full two-phase case.
The hysteresis and conservative regimes are well-captured, but the NODE-based formulation only approximates the sharp transitions in the data in part due to the coarseness of the time discretization (only 100 steps were used per trajectory).
A recent improvement in the ISV-NODE model \cite{jones2025attention} may enable the required abrupt transitions, but this application is left for future work.

\subsection{Waddington-inspired conservative gene network}

Waddington's epigenetic landscape is a classical metaphor in developmental biology \cite{baedke2013epigenetic}, where a particle following a potential surface represents a cell differentiating into one of several possible fates.
This metaphor can be translated into a dynamical systems framework using gene circuit models.
Consider two genes, $x_1$ and $x_2$, which regulate each other according to the following system:
\begin{equation}
\dot{\mathbf{x}} =
\begin{bmatrix}
\dot{x}_1\\\dot{x}_2
\end{bmatrix}=
\begin{bmatrix}
a x_1(p_1 -  x_1) + \dfrac{b}{S^n + x_2^n}
- \dfrac{b n x_1^{n-1} x_2}{(S^n + x_1^n)^2} \\[10pt]
a x_2(p_2  - x_2) + \dfrac{b}{S^n + x_1^n}
- \dfrac{b n x_2^{n-1} x_1}{(S^n + x_2^n)^2}
\end{bmatrix}
\label{eq:xdot_waddington}
\end{equation}

The first term in each component, $a x_i(p_i - x_i)$ for $i=1,2$, corresponds to a self-activation term with logistic dynamics, where $a$ controls the growth rate and $p_i$ sets the preferred expression level.
The second term, $b(S^n + x_j^n)^{-1}$ ($j \neq i$), represents a production term for $x_i$ that is suppressed by increasing levels of $x_j$, modeling cross-inhibition.
The final term in each equation captures nonlinear degradation due to the opposing gene, which becomes saturated in the limit of large $x_i$ but is linearly scaled by $x_j$.

This coupling structure induces bistability or multistability depending on the parameter regime.
For small $b$, self-activation dominates and the system favors a symmetric stable state near $(p_1, p_2)$.
For larger $b$, mutual inhibition dominates and leads to symmetry breaking, favoring differentiated states where one gene is highly expressed and the other is suppressed.

Although this model is synthetic, it is biologically motivated by the gene circuit studied by Wang et al.~\cite{wang2011quantifying}, where $x_1$ and $x_2$ represent the transcription factors PU.1 and GATA1, respectively.
In that model, each gene self-activates and represses the other, encoding a binary fate decision in hematopoietic stem cells.

Unlike the system in~\cref{wang2011quantifying}, our formulation admits a scalar potential $\Xi(x_1, x_2)$ such that the dynamics satisfy $\dot{\mathbf{x}} = -\nabla \Xi$.
This guarantees that the system is conservative and provides a direct interpretation in terms of an energy landscape, in the spirit of Waddington's original metaphor.

\begin{figure}
\centering
\begin{subfigure}[c]{0.45\textwidth}
\centering
\includegraphics[width=0.95\linewidth]{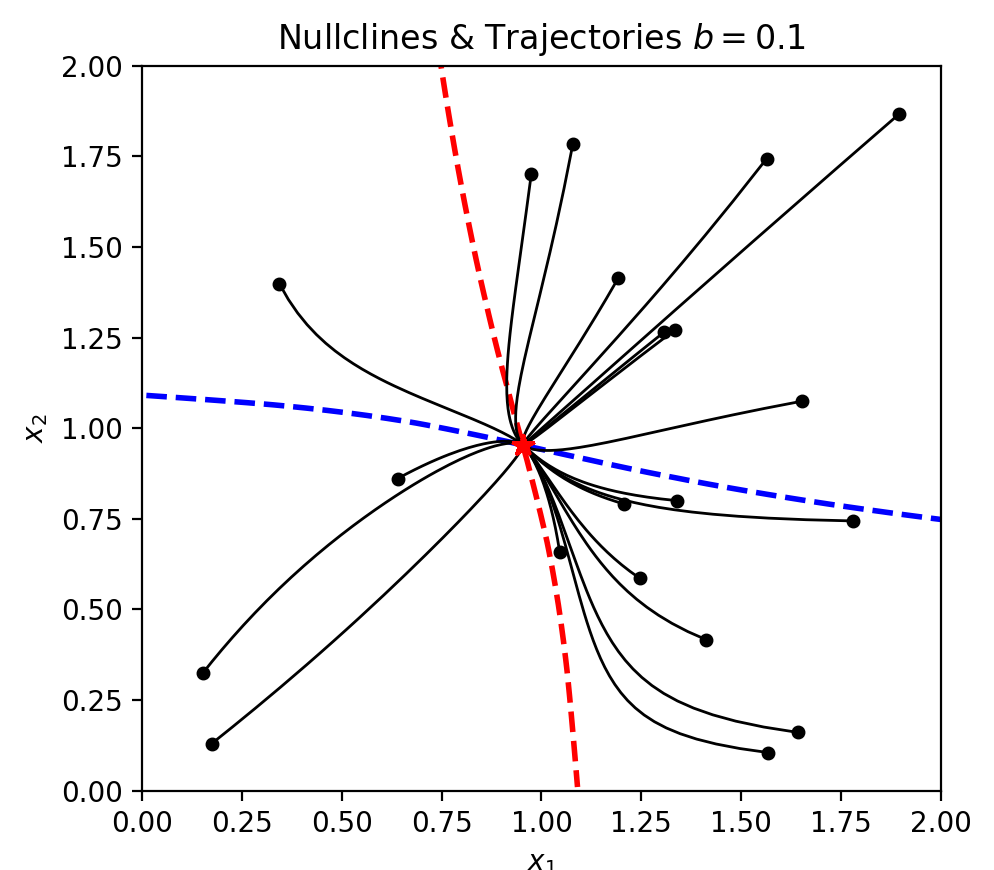}
\caption{}
\end{subfigure}
\begin{subfigure}[c]{0.45\textwidth}
\centering
\includegraphics[width=0.95\linewidth]{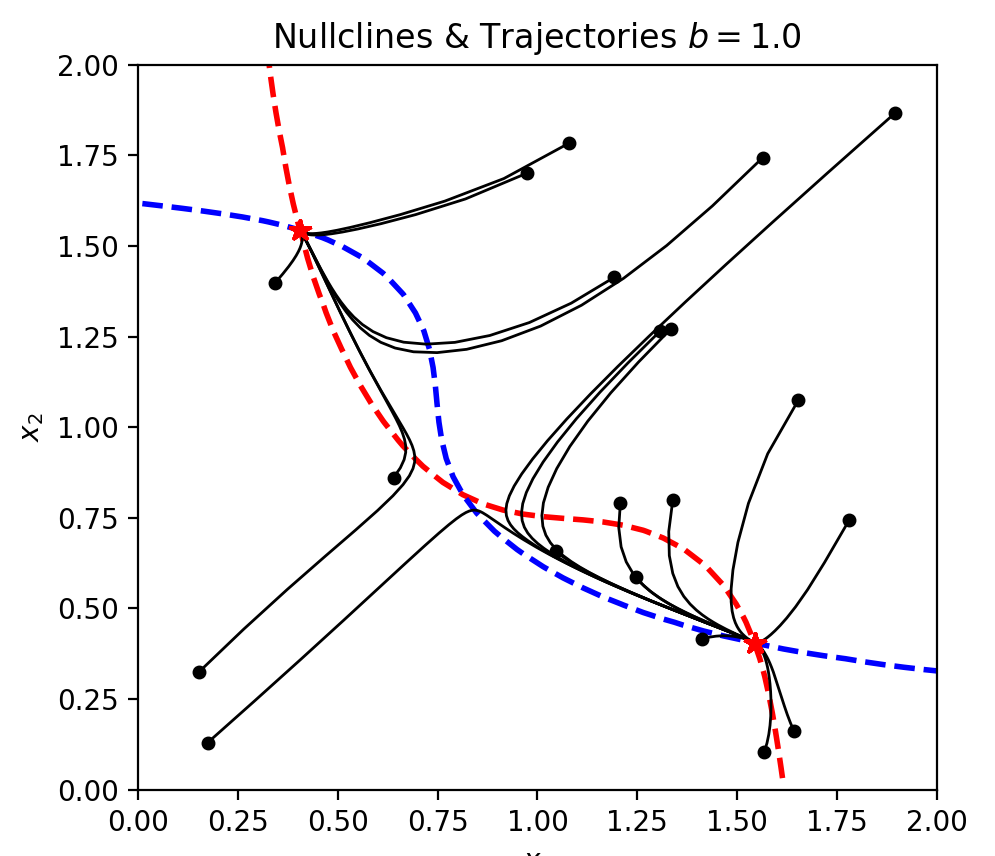}
\caption{}
\end{subfigure}
\begin{subfigure}[c]{0.45\textwidth}
\centering
\includegraphics[width=0.95\linewidth]{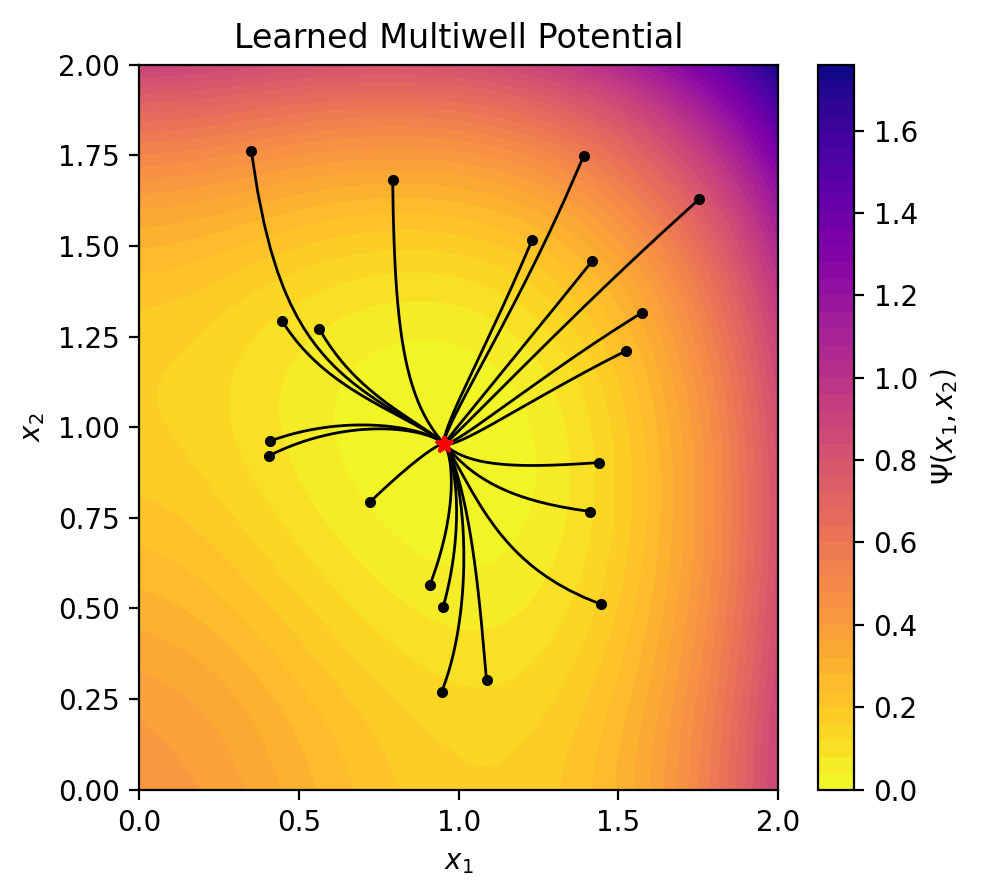}
\caption{}
\end{subfigure}
\begin{subfigure}[c]{0.45\textwidth}
\centering
\includegraphics[width=0.95\linewidth]{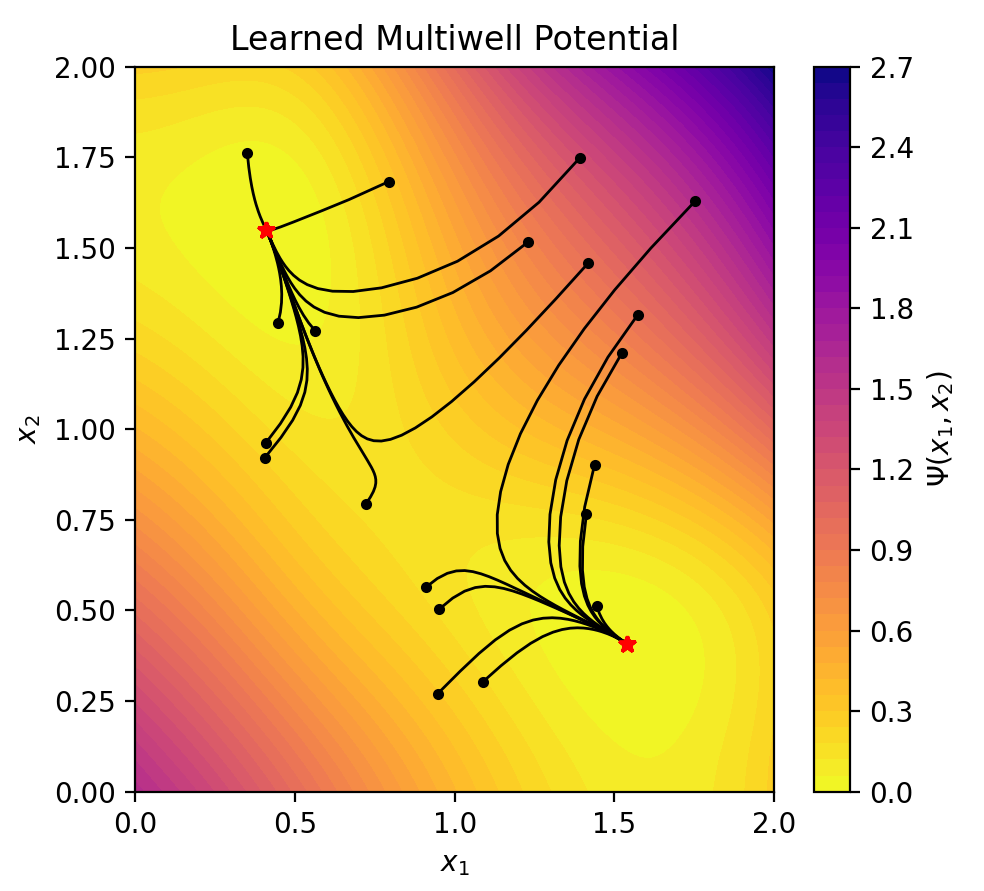}
\caption{}
\end{subfigure}
\caption{Gene trajectories and learned multi-well potentials for varying degree of cross-inhibition.}
\label{fig:waddington_learned_LSE}
\end{figure}

We trained two LSE-ICNN potentials independently, for the cases $b=0.1$ and $b=1.0$, respectively.
\fref{fig:waddington_learned_LSE}a shows trajectories from system
\eref{eq:xdot_waddington} for $b=0.1$ on top of the nullclines over the $(x_1,x_2)$ space.
For the small cross-inhibition, there is a single stable steady state at $\approx(1,1)$.
For large cross-inhibition $b=1$,
\fref{fig:waddington_learned_LSE}b shows trajectories on  $(x_1,x_2)$ space.
Depending on the initial condition, the system equilibrates at low-high $(x_1,x_2)$ or vice versa.
The nullclines explain the bistability of the system.

We picked LSE-ICNNs with 3 modes, and each ICNN mode had 3 hidden layers with 8 neurons each.
The trained multi-well potentials inferred from the dynamical system are illustrated in
\fref{fig:waddington_learned_LSE}c,d.
For $b=0.1$, a single dominant well for the potential $\Psi_1(x_1,x_2)$ is found such that the trajectories $\dot{\mathbf{x}}=-\nabla \Psi$ converge at a stable point $\approx(1,1)$.
For $b=1.0$, the learned potential $\Psi_2(x_1,x_2)$ discovers two wells such that the trajectories  $\dot{\mathbf{x}}=-\nabla \Psi$ approach one of the two wells.

\begin{figure}
\centering
\includegraphics[width=0.8\linewidth]{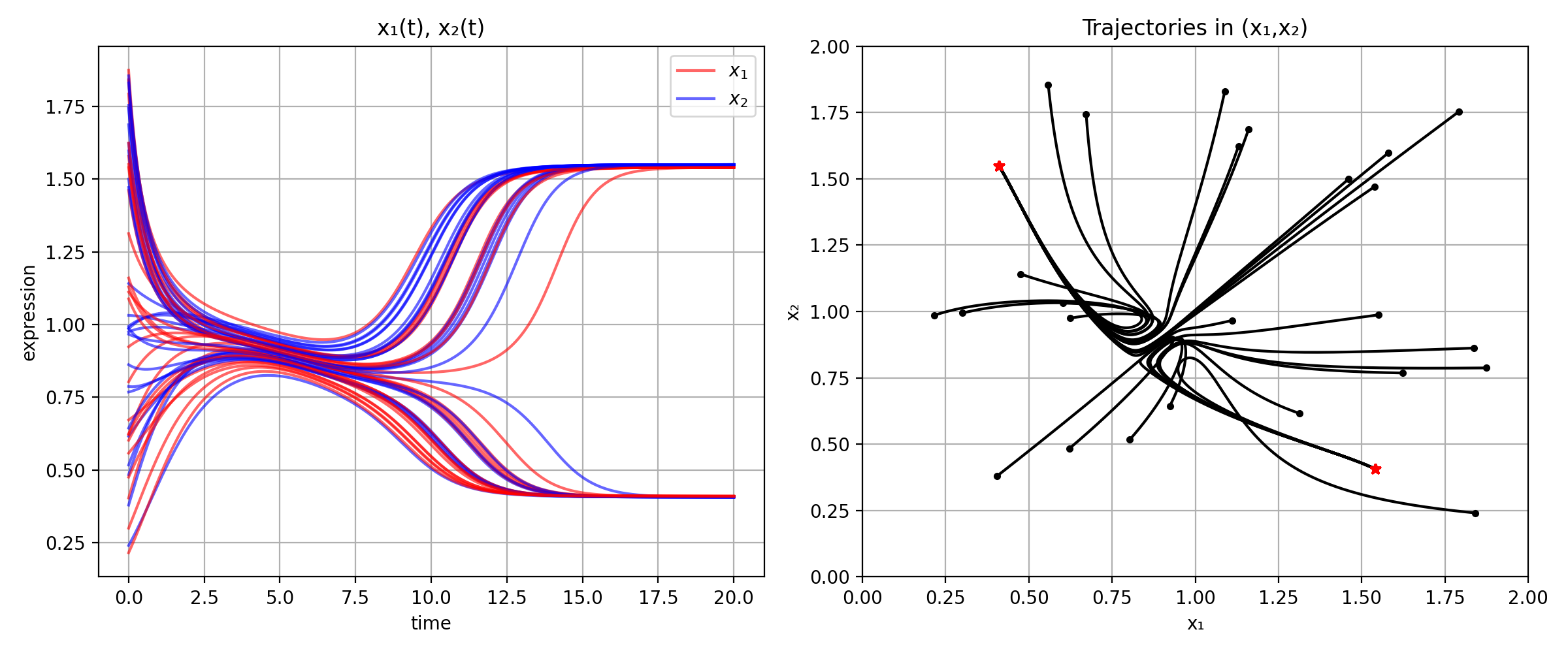}
\caption{Gene trajectories for a transient potential landscape that goes from a single-well to a double-well potential.}
\label{fig:waddington_transient_LSE}
\end{figure}

In \fref{fig:waddington_transient_LSE}, we create a transient LSE-ICNN potential through $\Psi_b = (1-b)\Psi_1+b\Psi_2$ with $\dot{b}=k_b(1-b)b$ and with $b(0)=0.01$, $k_b=0.7$.
The system trajectories are first attracted to the central well of $\Psi_1$ but eventually diverge to the two stable attractors of $\Psi_2$.

\section{Conclusion}\label{sec:conclusion}

In this work, we employ ICNNs as convex basis functions inside the LSE formulation, enabling a fully data-driven approximation of well-behaved multi-well energies.
We demonstrated that the LSE-ICNN is an effective representation across a variety of disciplines in computational physics.

Furthermore, we introduce a learnable transition scale and gating parameters associated with each well.
The inverse temperature-like parameter allows the LSE-ICNN to fit the range of smooth to sharp transitions.
The gating weights act as an adaptive mechanism that determines the relevance of each mode during training, effectively allowing the model to infer the number of active wells from data.
This feature is particularly valuable in situations where the number of energy minima is not known \apriori.

Overall, our method provides a new paradigm for learning energy functions that are differentiable, convex within basins, and adaptive to complex multimodal structures such as: multimodal probability densities, loss of polyconvexity in materials with phase-changes or instabilities during microstructure deformation, and biological networks with multiple stable wells guiding cell differentiation.

In future work, we intend to continue the development of the LSE-ICNN in dynamic and dissipative applications, which include further thermodynamic considerations to guarantee positive energy dissipation. By bridging physical priors with flexible neural architectures, the LSE-ICNN opens new possibilities for robust, data-driven modeling of complex energy landscapes in computational physics.

\section*{Acknowledgments}

Sandia National Laboratories is a multimission laboratory managed and operated by National Technology and Engineering Solutions of Sandia, LLC., a wholly owned subsidiary of Honeywell International, Inc., for the U.S.
Department of Energy's National Nuclear Security Administration under contract DE-NA-0003525.
This paper describes objective technical results and analysis.
Any subjective views or opinions that might be expressed in the paper do not necessarily represent the views of the U.S.  Department of Energy or the United States Government.



\appendix
\raggedbottom
\setcounter{equation}{0}
\renewcommand{\theequation}{A-\arabic{equation}}
\section{Elastic stability model} \label{app:elastic_stability}

For a two-phase material Rossi \etal \cite{rossi2024limit} assume a mixture free energy function of strain $\strain$ and internal variable $c$ characterizing phase fraction:
\begin{equation}
\Psi(\strain,c) = (1-c) \psi_A(\strain_A) + c \psi_B (\strain_B)
\end{equation}
where $\strain = (1-c) \strain_A + c \strain_B$ and
\begin{equation}
\partialb_\strain \Psi_A(\strain_A) =
\partialb_\strain \Psi_B(\strain_B)
\end{equation}
in equilibrium.
The particular form of the individual phase free energy:
\begin{equation}
\psi(\strain) = \frac{1}{2} (\strain - \strain_0) \Cbb  (\strain - \strain_0)  + \Psi_0
\end{equation}
is consistent with linear elasticity with $\Cbb$ being the elastic modulus tensor for the phase.
As in \cref{rossi2024limit} we employ an isotropic $\Cbb = K \Ib \otimes \Ib + 2 G (\Ibb - \tfrac{1}{3} \Ib\otimes\Ib)$ with the same bulk modulus $K$ and shear modulus $G$ for both.
Without loss of generality, we can assume that the reference energy for A is zero:
$\psi_0 = 0$, so that the difference in the reference energy $\Delta \psi$ is the relevant parameter.
Similarly, we assume $\strain_0 = \mathbf{0}$ for the A phase, and  $\strain_0 = \epsilon_0 (\eb_1 \otimes \eb_1)$ for the B phase.
So that the free energy is given by
\begin{equation}
\Psi(\strain,c) = \frac{1}{2} (1-c) \strain_A : \Cbb  \strain_A + \frac{1}{2} c (\strain_B - \strain_0) : \Cbb  (\strain_B - \strain_0)  +  c \Delta\Psi \ ,
\end{equation}
the stress by
\begin{equation}
\stress = \partialb_\strain \Psi
= \Cbb (\strain - c \strain_0)
\end{equation}
and the chemical potential (conjugate force to $c$) by:
\begin{equation}
\kappa = -\partial_c \Psi
= \stress \cdot \strain_0 - (\psi_B - \psi_A) .
\end{equation}
The mechanical dissipation $\Gamma$ due to change of state $c$ is simply:
\begin{equation}
\stress \cdot \dot{\strain} - \dot{\Psi}
= \kappa \dot{c} \ge 0.
\end{equation}

The dissipation potential in this generalized standard material framework \cite{halphen1975materiaux,ziegler1987derivation}:
\begin{equation}
\phi(c,\dot{c})
=
\begin{cases}
+k_{+}\, \dot{c}, & \text{if}  \, \, \dot{c} \ge 0 \ \text{and} \ c \in [0,1),   \\
-k_{-}\,  \dot{c}, & \text{if}  \, \,  \dot{c} < 0 \ \text{and} \ c \in (0,1],  \\
\infty, & \text{else}.
\end{cases}
\end{equation}
is not smooth due to the constraint $c \in [0,1]$; hence the flow rule is:
\begin{equation}
\dot{c}
=
\begin{cases}
+k_{+}, &  \text{if} \, \,   \kappa = +k_+,   \\
-k_{-}, & \text{if} \, \,   \kappa = -k_- ,   \\
0, &\text{else} .
\end{cases}
\end{equation}
The Biot relation
\begin{equation}
\kappa \equiv \stress\cdot\strain_0 - \Delta \psi =
\partial_{\dot{c}} \phi
= \begin{cases}
+k_+ &\\
-k_- &
\end{cases}
\end{equation}
can be solved so the stress is bounded when $c \in (0,1)$:
\begin{equation}
\strain_0 \Cbb (\strain - c \strain_0) - \Delta \psi =
- c \strain_0 \Cbb \strain_0
+ \strain_0 \Cbb \strain
- \Delta \psi
=  k_\pm
\end{equation}
which leads to:
\begin{equation}
c = \frac{1}{\strain_0 \Cbb \strain_0}
(\strain_0 \Cbb \strain - \Delta \psi  - k_\pm)  \ .
\end{equation}
\end{document}